\documentclass{article}

\usepackage{microtype}
\usepackage{graphicx}
\usepackage{subcaption}
\usepackage{booktabs} 

\usepackage{hyperref}

\usepackage[preprint]{icml2026}

\usepackage{amsmath}
\usepackage{amssymb}
\usepackage{mathtools}
\usepackage{amsthm}

\usepackage{bm}  
\usepackage{amsmath}  
\usepackage{amsthm}  
\usepackage{amssymb}  
\usepackage{graphicx}  
\usepackage{algorithm}  
\usepackage{algorithmic,enumitem}  
\usepackage{subcaption} 
\usepackage{multirow} 
\usepackage{booktabs}
\usepackage{xcolor} 
\usepackage[normalem]{ulem}

\usepackage[capitalize,noabbrev]{cleveref}

\theoremstyle{plain}
\newtheorem{theorem}{Theorem}[section]

\newtheorem{lemma}[theorem]{Lemma}

\theoremstyle{definition}

\theoremstyle{remark}
\newtheorem{remark}[theorem]{Remark}

\usepackage[textsize=tiny]{todonotes}

\icmltitlerunning{Diffusion Models with Double Guidance}

\begin{document}

\twocolumn[
  \icmltitle{Diffusion Models with Double Guidance: Generate with aggregated datasets}



  \icmlsetsymbol{equal}{*}

  \begin{icmlauthorlist}
    \icmlauthor{Yanfeng Yang}{sokendai,ism}
    \icmlauthor{Kenji Fukumizu}{ism}
  \end{icmlauthorlist}

  \icmlaffiliation{sokendai}{Graduate University of Advanced Studies (SOKENDAI), Kanagawa, Japan}
  \icmlaffiliation{ism}{The Institute of Statistical Mathematics, Tokyo, Japan}

  \icmlcorrespondingauthor{Yanfeng Yang}{yanfengyang0316@gmail.com}
  \icmlcorrespondingauthor{Kenji Fukumizu}{fukumizu@ism.ac.jp}

  \icmlkeywords{Machine Learning, ICML}

  \vskip 0.3in
]



\printAffiliationsAndNotice{}  

\begin{abstract}
  Creating large-scale datasets for training high-performance generative models is often prohibitively expensive, especially when associated attributes or annotations must be provided. As a result, merging existing datasets has become a common strategy. However, the sets of attributes across datasets are often inconsistent, and their naive concatenation typically leads to block-wise missing conditions. This presents a significant challenge for conditional generative modeling when the multiple attributes are used jointly as conditions, thereby limiting the model’s controllability and applicability. To address this issue, we propose a novel generative approach, Diffusion Model with Double Guidance, which enables precise conditional generation even when no training samples contain all conditions simultaneously. Our method maintains rigorous control over multiple conditions without requiring joint annotations. We demonstrate its effectiveness in molecular and image generation tasks, where it outperforms existing baselines both in alignment with target conditional distributions and in controllability under missing condition settings.
\end{abstract}

\section{Introduction}
Diffusion-based generative models have achieved remarkable success across a range of data modalities, including images, videos, molecules, and text \cite{ho_ddpm,song_sgm,Rombach_2021_latent_diffusion,ho_2022_video,hoogeboom_2022_e3_diffusion,Minkai_2025_textdiffusion}. In many real-world applications, the goal is not simple generation from the unconditional distribution $P_{X_0}$ of data $X_0\in \mathbb{R}^d$, but rather from the conditional distribution $P_{X_0|C}$ given some auxiliary conditions $C \in \mathbb{R}^p$ such as labels and properties.  Conditional diffusion models (CDMs) enable such generation under conditioning information, allowing fine-grained control over content. In computer vision, CDMs generate images conditioned on object categories, spatial layouts, or styles \cite{Song_2023_lgd_guidance,zhang_2023_control_net,zhao_2025_adding_cond_by_rl,chung_diffusion_inverse,ho_classifier_free,zhao_finetune_cm,dhariwal_guided_diffusion}. In medical imaging, CDMs have been used to rectify tilted CT images \cite{kawata_2025_multimodal_diffusion}. In chemistry, CDMs enable the generation of molecules with the desired physicochemical properties, facilitating drug discovery and material design \cite{Gebauer_2022_cond_drug_design_1,SY_chemguide,Ninniri_2023_cfg_graph_molecular}. In time series, CDMs provide effective approaches to probabilistic forescasting \cite{ye_2025_nsdiff, yang_2025_cwgen}. Beyond these domains, CDMs have also been applied to Bayesian inference and causal discovery \cite{gloeckler_2025_markovsbi,yang_2024_cdcit,Sanchez_2022_diffusion_causal}.

Despite the strong expressive power of CDMs, training a high-quality CDM critically depends on the sample size of the training data \cite{chen_2023_sampling_as_easy, Fu_2024_UnveilCD}. 
A common strategy to increase the amount of training data is to merge existing datasets into an aggregated one~\cite{Kaufman_2024_coatildm,google_2022_image_concat_datasets,Li_2023_diffusion_models_sample_size,Fu_2024_UnveilCD}. However, datasets collected from different sources often contain distinct types of conditioning information. In the simplified setting illustrated in Table~\ref{table_block_wise_missing}, dataset $D^{(1)}$ contains no samples labeled with condition $C_2$, while dataset $D^{(2)}$ lacks samples associated with condition $C_1$. 
This naturally gives rise to a fundamental challenge: how can a CDM capture the conditional distribution with joint conditions $P_{X_0|C_1,C_2}$ with aggregated datasets containing block-wise missing conditions? Addressing this challenge would substantially enhance the controllability of diffusion models and broaden the range of tasks they can support.

As a motivating example, molecular generation tasks in chemistry exhibit particularly pronounced challenges arising from data aggregation. In drug discovery, conditional generative models have been widely used to design molecular candidates that exhibit the desired physicochemical properties~\cite{Haote_Li_KAE,SY_chemguide,hoogeboom_2022_e3_diffusion,Gebauer_2022_cond_drug_design_1}. However, constructing large-scale molecular datasets annotated with additional properties remains a formidable challenge in terms of both cost and time. Laboratory measurements are often prohibitively expensive~\cite{Erhirhie_2018_hard_to_get_lethal_dose,SY_chemguide}, while the accurate computation of molecular properties is notoriously time-consuming~\cite{Fernandez_2024_xray_molecular,Jarrold_2022_mass_spec_molecular}.

Conventional approaches to accessing the combined conditions have been considered. A direct approach is to recover missing conditional information via manual experiments or imputation approaches. As discussed above, the former is often impractical; the latter, imputation, typically relies on strong correlations between variables~\cite{Yoon_2018_gain,Jolicoeur_2024_forestdiffusion}. However, in our case, the datasets may lack samples drawn from the joint distribution $P_{C_1, C_2}$, which severely compromises the performance of standard imputation methods, leading to inaccurate results. 

Recent generative approaches to approximating $P_{X_0|C_1,C_2}$ from block-wise missing datasets include composing different score functions \cite{Liu_2022_compositional_diff,gaudi_2025_coind}. In parallel, other strategies leverage ControlNets \cite{zhang_2023_control_net} to encode additional conditions into a pre-trained conditional generative model, with further fine-tuning via reinforcement learning (RL) \cite{zhao_2025_adding_cond_by_rl}. Although compositional methods offer a flexible way to combine multiple conditions, their theoretical foundation may not have rigor \cite{bradley_2025_Composition}. On the other hand, the RL-based method relies on a strong pre-trained diffusion model \cite{zhao_2025_adding_cond_by_rl}, and the reward function neglects finer-grained control over the intermediate steps of the reverse process \cite{Zijing_2025_rl_is_bad}.

In light of these limitations, we argue that an approach should be both explainable and training-free. Moreover, it should leverage the dependence structure among variables, without relying on additional experiments or imputations.

In this work, we assume access to a pre-trained diffusion model that has already captured the distributions $P_{X_0}$ and $P_{X_0|C_1}$. Building on this foundation, we develop methods to generate samples from the joint conditional distribution $P_{X_0|C_1,C_2}$, even in the absence of joint samples from $P_{C_1,C_2}$ being available. Our approach leverages the assumption that $C_1$ and $C_2$ are conditionally independent given $X_0$ and exploits this structure to introduce an additional guidance term into the generation process.

Our main contributions are summarized as follows.
\begin{itemize}
    \item \textbf{Plug-and-play conditional generation on block-wise missing datasets:}  
    We propose two efficient methods for generating samples approximately from $P_{X_0|C_1,C_2}$ when we only have access to block-wise missing datasets shown in Table \ref{table_block_wise_missing}.  
    \item \textbf{Theoretical analysis and enhanced interpretability for guided diffusion:} We establish a more detailed theoretical framework compared to previous works, providing a deeper understanding of the factors that influence the effectiveness of guided diffusion.
    \item \textbf{Extensive experiments across multiple data types:} We evaluate our methods on synthetic datasets, molecular datasets \cite{Axelrod_2022_geomdataset,Sterling_2015_zincdataset}, and image datasets \cite{Deng_2009_imagenet}, demonstrating superior recovery of conditional distributions and controllability over the target conditions. 
\end{itemize}

The remainder of this paper is organized as follows. Section~\ref{sec_Preliminary} introduces the problem setting and background. Section~\ref{sec_dmdg} describes our proposed methods. 
Section~\ref{sec_thm} presents the theoretical analysis. Section~\ref{sec_mol} includes two practical applications to molecular datasets.  Section \ref{sec_image_gen} presents image inpainting and object editing. Section \ref{sec_conclusion} summarizes our results and provides a discussion of future directions.

\begin{table}[htbp!]
\centering
\caption{An illustration of aggregated datasets. Concatenating them makes a block-wise missing dataset. Missing values are represented as $\varnothing$.}
\label{table_block_wise_missing}
\centering
\begin{tabular}{|c|c|c|c|}
\hline
Dataset & Target variable & Condition 1 & Condition 2 \\ \hline
$D^{(1)}$ & $X_0^{(1)}$ & $C_1^{(1)}$ & $\varnothing$ \\ \hline
$D^{(2)}$ & $X_0^{(2)}$ & $\varnothing$ & $C_2^{(2)}$ \\ \hline
\end{tabular}
\end{table}

\section{Problem Setting and Preliminaries} 
\label{sec_Preliminary}

\subsection{Problem setting}
\label{sec_problem_set}
Our algorithms are designed to operate on function-mapping datasets, which consist of samples $(X_0, C)$. The condition $C$ can be decomposed into two distinct parts $(C_1,C_2)$ by dimension, where $C_1 \in \mathbb{R}^k$ and $C_2 \in \mathbb{R}^{p-k}$. Both $C_1$ and $C_2$ can be expressed by two deterministic functions $f_1: \mathbb{R}^d \to \mathbb{R}^{k}$ and $f_2: \mathbb{R}^d \to \mathbb{R}^{p-k}$ with additive noise:
\begin{align}
\label{eq_function_map}
 \begin{pmatrix}
C_1 \\
C_2
\end{pmatrix}
&=
\begin{pmatrix}
f_1(X_0) \\
f_2(X_0) 
\end{pmatrix}
+
\begin{pmatrix}
\epsilon_1 \\
\epsilon_2
\end{pmatrix}, 
\end{align}
where $\epsilon_1 \sim N(0,\sigma_1^2 I_k), \epsilon_2 \sim N(0,\sigma_2^2 I_{p-k}), \epsilon_1 \perp \! \! \! \perp \epsilon_2$, and $\sigma_1, \sigma_2 >0$.
An important consequence of this assumption is the
{\bf conditional independence} (CI), denoted by $C_1 \perp \!\!\! \perp C_2 | X_0$.  Note that this is different from (unconditional) independence $C_1 {\perp \!\!\! \perp} C_2$; even if they are correlated, the conditional independence given 
$X_0$ may hold. 

Function-mapping datasets are ubiquitous and naturally arise from a variety of domains of science. Molecular datasets, for example, are typical function-mapping datasets \cite{Axelrod_2022_geomdataset,Sterling_2015_zincdataset,ramakrishnan_2014_qm9dataset}, where $X_0$ represents the structure of a molecule and $C_1,C_2$ correspond to different properties of the molecule. Note that in this case, two properties $C_1$ and $C_2$ can be correlated; while given a molecule $X_0$, the properties can be deterministic and therefore conditionally independent.  Similarly, for an image $X_0$ of an object, the condition $C_1$ may represent the category of the object, while $C_2$ is the transformed $X_0$, such as masked image or down-sampling \cite{chung_diffusion_inverse,gaudi_2025_coind,Liu_2022_compositional_diff}.


Although different datasets consistently include samples of $X_0$, the available conditions can vary due to technical constraints or limitations of the data acquisition devices (Table \ref{table_block_wise_missing}). 
For ease of exposition, we mainly discuss the case of 
two datasets with partially observed conditions: $D^{(1)} = \{ X_0^{(1)}, C_1^{(1)}, \varnothing \}$ and $D^{(2)}= \{X_0^{(2)}, \varnothing, C_2^{(2)} \}$. More general cases involving more than two 
datasets and missing conditions can be found in Appendix \ref{sec_dmtg}. 
Additionally, in practical situations, 
the empirical distributions of $X_0^{(1)}$ and $X_0^{(2)}$ 
may not be exactly the same but only be similar.  
More detailed discussion of such potential distribution shifts is discussed in Appendix \ref{sec_app_th_bound_risk}.

\subsection{EDM framework}
\label{sec_edm}
Diffusion models consist of a forward and a reverse process. In the forward process, Gaussian noises are gradually added to the original data, and the score function of the noisy data is learned. 
The reverse process then employs stochastic differential equations (SDE) or ordinary differential equations (ODE) with the learned score function to convert a Gaussian noise into a sample that closely approximates the original data distribution \cite{ho_ddpm,song_sgm}. We employ the framework of Elucidated Diffusion Model  \citep[EDM,][]{karras_edm}, which represents a streamlined diffusion model using the simple and intuitive forward process:
\begin{equation}
\label{eq_forward_process}
    dX_t=\sqrt{2t}\,dB_t, \quad t \in [0,t_{\max}],
\end{equation}
where $B_t$ is a Wiener process. By the property of Ornstein-Uhlenbeck process, we have $X_t \overset{d}{=} X_0 + t\epsilon, \epsilon \sim N(0,I_d)$ \cite{yang_2024_cdcit}. Analytically, the ODEs used for unconditional and conditional sampling in EDM's reverse process are given by, respectively,:
\begin{align}
\label{eq_edm_reverse}
    dX_t & =-t \cdot \nabla  \log p_t(X_t) dt, \notag \\
    dX_t & =-t \cdot \nabla  \log p_t(X_t|C_1) dt, \quad t \in [t_{\min},t_{\max}],
\end{align}
where $p_t(X_t)$ is the density of $X_t$ and $p_t(X_t|C_1)$ is the conditional density of $X_t$ given $C_1$. In this paper, unless otherwise specified, $\nabla$ denotes the gradient with respect to $X_t$. To avoid numerical explosions, the reverse process uses an early-stopping at $t_{\min}$ close to but larger than zero. Since the score functions $\nabla \log p_t(X_t)$ and $\nabla  \log p_t(X_t|C_1)$ lack explicit analytical expressions, we train a neural network $nn_\theta$, parameterized by $\theta$, to approximate them simultaneously by minimizing the following loss:
\begin{equation}
\label{eq_edm_loss}
    \mathcal{L}=
    \begin{cases}
    \mathbb{E}_{X_0,t,\epsilon} \|  X_0 - nn_{\theta} \left(X_0 +t \epsilon, t, \varnothing \right) \|_2^2, \quad \quad \ \    p_{\text{non}} \\
    \mathbb{E}_{X_0,C_1,t,\epsilon} \| X_0 - nn_{\theta} \left(X_0 + t\epsilon, t,C_1 \right) \|_2^2, \    1 - p_{\text{non}}
    \end{cases}
\end{equation}
where $\log t \sim N(-1.2,1.2^2)$, $X_0 \sim P_{X_0}$, $(X_0,C_1) \sim P_{X_0,C_1}$ and $p_{\text{non}} \in [0,1]$ is a predefined parameter controlling the probability of masking the condition $C_1$. The unconditional and conditional score functions are estimated as $s_{\theta}(X_t, t, \varnothing) := [nn_{\theta}(X_t, t, \varnothing) - X_t]/t^2$ and $s_{\theta}(X_t, t, C_1) := [nn_{\theta}(X_t, t, C_1) - X_t]/t^2$, respectively. These estimators enable the reverse process to generate samples that approximately follow $P_{X_0}$ and $P_{X_0 | C_1}$. By employing an ODE solver, EDM requires substantially fewer sampling steps than SDE-based methods \cite{ho_ddpm, song_sgm}, enabling faster generation.

\subsection{Classifier Guidance (CG) and Diffusion Posterior Sampling (DPS)}

\citet{dhariwal_guided_diffusion} proposed classifier-guided diffusion, a training-free method to add conditional control to an unconditional diffusion model. Using  Bayes' rule, the conditional score function admits the decomposition:
\begin{equation}
\label{eq_cond_score_true}
    \nabla  \log p_t(X_t|C_1) =  \nabla  \log p_t(X_t) + \nabla \log p(C_1|X_t),
\end{equation}
where $ p(C_1|X_t)$ is the density of $C_1$ given noisy $X_t$. In this paper, we mainly discuss the case in which conditions $C_1$ and $C_2$ are continuous. Let $\widetilde{f}_1(X_t)$ approximate $\mathbb{E}[C_1 | X_t]$.
The density $p(C_1 | X_t)$ is approximated by a Gaussian
$N(\widetilde{f}_1(X_t), (2\lambda_1)^{-1} I_k)$,
which yields the tractable form
$\nabla \log p(C_1 | X_t) \approx -\lambda_1 \nabla \| C_1 - \widetilde{f}_1(X_t) \|_2^2$,
known as classifier guidance (CG). Incorporating CG into (\ref{eq_cond_score_true}) yields:
\begin{equation}
\label{eq_cond_score}
    \nabla \log p_t(X_t|C_1) \approx s_{\theta}(X_t, t, \varnothing) - \lambda_1 \nabla  \| C_1 - \widetilde{f}_1(X_t) \|_2^2.
\end{equation}
Here, $\lambda_1$, referred to as the guidance scale, is a hyperparameter that controls the strength of condition during generation. Its choice depends on the specific task and reflects a trade-off between conditional fidelity and sample quality: larger values of $\lambda_1$ enforce stronger adherence to the condition $C_1$ but may degrade sample quality.

When $X_0$ and $C_1$ satisfy the function-mapping relationship in (\ref{eq_function_map}), \citet{chung_diffusion_inverse} proposed Diffusion Posterior Sampling (DPS). In DPS, the mapping function $f_1$ is assumed to be known and operates on the posterior mean $\mathbb{E}\left[ X_0 | X_t \right]$ obtained via Tweedie projection:
\begin{equation}
\label{eq_tweedie_proj}
    \mathbb{E}\left[ X_0 | X_t \right]=X_t+t^2 \cdot \nabla \log p_t(X_t).
\end{equation}
By replacing the intractable score function in~\eqref{eq_tweedie_proj} with its estimator $s_{\theta}(X_t, t, \varnothing)$, the posterior mean $\mathbb{E}[X_0 | X_t]$ is estimated as $X_{0 | t} := nn_{\theta}(X_t, t, \varnothing)$. Accordingly, the gradient $\nabla \log p(C_1 | X_t)$ is approximated by $- \lambda_1 \nabla \| C_1 - f_1(X_{0 | t}) \|_2^2$. Even when $f_1$ is unknown, it can be approximated by a regressor $\widehat{f}_1(X_0)$; the estimator $- \lambda_1 \nabla \| C_1 - \widehat{f}_1(X_{0 | t}) \|_2^2$ is typically more accurate than $-\lambda_1 \nabla \| C_1 - \widetilde{f}_1(X_t) \|_2^2$ in \eqref{eq_cond_score}. As a result, conditional samples generated by DPS often exhibit higher quality than those obtained via CG~(\ref{eq_cond_score}).

\begin{figure*}[htbp!]
    \centering
    \includegraphics[width=0.9\linewidth]{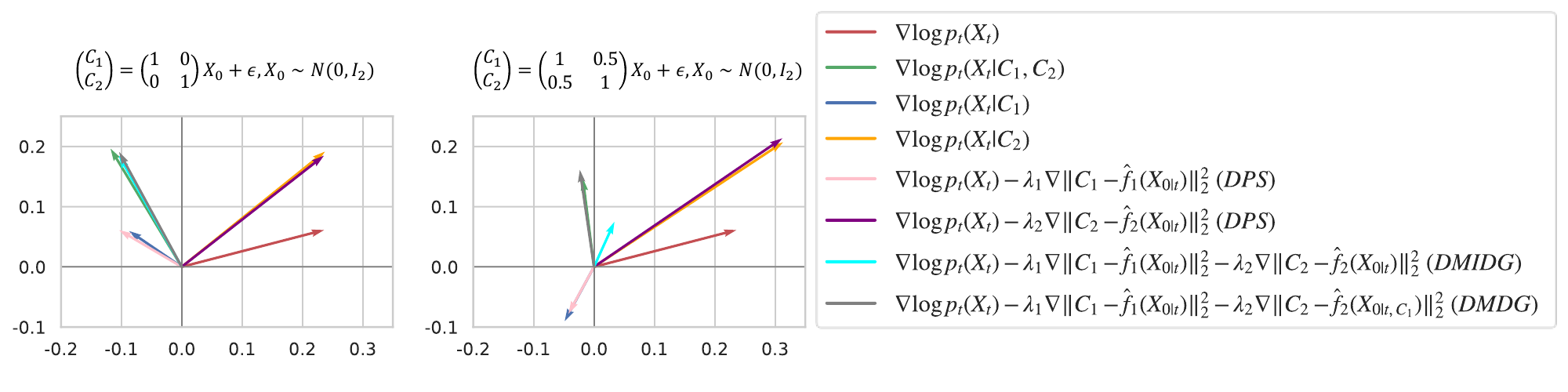}
    \caption{Comparison of score functions when $t=1$. In the left panel, $C_1 \perp \! \! \! \perp C_2$, and both DMDG (proposed) and DMIDG (previous) can accurately approximate the conditional score function $\nabla \log p_t(X_t|C_1,C_2)$.  In the right panel, $C_1 \not \perp \! \! \! \perp C_2$, and DMIDG can no longer approximate the correct score function, whereas DMDG remains effective. The error of DMIDG comes from the wrong assumption $C_1 \perp\!\!\!\perp C_2 | X_t$.}
    \label{fig_vector_fields}
\end{figure*}

\subsection{Classifier-Free Guidance (CFG)}
\label{sec_cfg}
Although inserting CG into an unconditional score allows conditional sampling with adjustable guidance scale, it can lead to a distorted conditional distribution with excessively complex $C_1$ \cite{ho_classifier_free} or improper guidance scale $\lambda_1$ \cite{Wu_2024_big_scale_distort_cond_dist}. To address this issue, \citet{ho_classifier_free} proposed classifier-free guidance (CFG), which achieves high-quality conditional sampling while retaining control over the conditioning strength. CFG replaces the second gradient term in (\ref{eq_cond_score_true}) with $\lambda_1 \bigl( \nabla \log p_t(X_t|C_1) - \nabla \log p_t(X_t) \bigr)$, resulting in:
\begin{equation*}
    \nabla \log p_t(X_t|C_1) \approx (1-\lambda_1) s_{\theta}(X_t, t, \varnothing) + \lambda_1 s_{\theta}(X_t, t, C_1).
\end{equation*} 
Compared to CG, CFG is capable of handling  more intricate conditions during the reverse process and typically generates higher-quality samples, particularly in image generation.

\section{Methods of Double Guidance}
\label{sec_dmdg}

This section first review the previous double guidance method and its theoretical flaw, then explains our proposed methods: Diffusion model with double guidance (DMDG) and Diffusion model with hybrid guidance (DMHG).

\subsection{Existing methods}
Generating samples from $P_{X_0 | C_1, C_2}$ via CG, DPS, or CFG requires either modeling the conditional densities $p(C_1, C_2 | X_t)$,  $p(C_1, C_2 | \mathbb{E}[X_0 | X_t])$ or the conditional score $\nabla \log p_t(X_t|C_1,C_2)$, respectively. However, as shown in Table~\ref{table_block_wise_missing}, the lack of joint samples of $(X_0, C_1, C_2)$ renders CG, DPS and CFG infeasible.

Prior to our work, \citet[][Appendix D.4]{ye_2024_unified_guidance} proposed to approximate $p(C_1, C_2 | X_t)$ by $p(C_1 | X_t) p(C_2 | X_t)$ which does not require joint samples of $(X_0, C_1, C_2)$. However, this approximation assumes
$C_1 \perp\!\!\!\perp C_2 | X_{\color{red} t}$, an assumption that does not hold in general. Even if $C_1 \perp\!\!\!\perp C_2 | X_{\color{red} 0}$ hold in general, the diffused $X_t$ may not give the conditional independence. As illustrated in Figure \ref{fig_vector_fields}, \ref{fig_CI_relationship} and Appendix \ref{sec_app_dmihg}, such an assumption can cause error in the estimated score function. We refer to this approach as Diffusion Model with \textit{Independent} Double Guidance (DMIDG), while the original authors term it \textit{combined guidance}. Detailed descriptions can be found in Appendix \ref{sec_app_dmihg}.

\subsection{Diffusion model with double guidance (DMDG)}

Motivated by the limitations of previous methods, we develop a novel guidance method to decompose the conditional score, avoiding the use of the joint samples of $(X_0,C_1,C_2)$.  Inspired by \eqref{eq_cond_score} and DPS, 
we decompose the conditional score function as:
\begin{align}
\label{eq_our_cond_score_main_tex}
    & \nabla \log p_t(X_t|C_1, C_2) \notag \\
    & = \nabla \log p_t(X_t) + \nabla \log p(C_1|X_t) + \nabla \log p(C_2|X_t,C_1) \notag \\
    & \approx \nabla \log p_t(X_t)  + \nabla \log p(C_1|\mathbb{E} \left[ X_0|X_t\right] ) \notag \\
    & \quad + \nabla  \log p(C_2|\mathbb{E}\left[X_0|X_t,C_1\right]),
\end{align}
where $p(C_1|\mathbb{E} \left[ X_0|X_t\right] )$ and $p(C_2|\mathbb{E}\left[X_0|X_t,C_1\right])$ are  the densities of  $N ( f_1(\mathbb{E} \left[ X_0|X_t\right]), \sigma_1^2 I_k)$ and $N(f_2(\mathbb{E}\left[X_0|X_t,C_1\right]), \sigma_2^2 I_{p-k})$, respectively, by the defination in (\ref{eq_function_map}). The posterior means $\mathbb{E} \left[ X_0|X_t\right]$ and $\mathbb{E}\left[X_0|X_t,C_1\right]$, obtained via Tweedie projection (\ref{eq_tweedie_proj}), are estimated by $X_{0|t}=nn_\theta(X_t,t, \varnothing)$ and $X_{0|t,C_1}:=nn_\theta(X_t,t,C_1)$, respectively. During the reverse process, we need to compute $f_1, f_2, \nabla f_1$ and $ \nabla f_2$ multiple times. When these functions are unknown, non-differentiable, or costly, we instead train neural regressors (or classifiers) $\widehat{f}_1$ and $\widehat{f}_2$ to approximate $f_1$ and $f_2$. Subsequently, similarly to (\ref{eq_cond_score}), $p(C_1|\mathbb{E} \left[ X_0|X_t\right] )$ and $p(C_2|\mathbb{E}[X_0|X_t, C_1])$ are approximated by $p(C_1|X_{0|t})$ and $p(C_2|X_{0|t,C_1})$, respectively. Equivalently, they are modeled by the following normal distributions:
\begin{equation}
\label{relationship_nerual_estimator}
\begin{pmatrix}
C_1 \\
C_2
\end{pmatrix}
 \sim N \left(
\begin{bmatrix}
\widehat{f}_1(X_{0|t}) \\
\widehat{f}_2(X_{0|t,C_1})
\end{bmatrix},
\begin{bmatrix}
\dfrac{1}{2\lambda_1}  I_k & 0 \\
0 & \dfrac{1}{2\lambda_2}  I_{p-k}
\end{bmatrix}
\right), 
\end{equation}
where $\lambda_1,\lambda_2$ are positive guidance scales used to adjust the strength of guidance. Finally, replacing $\nabla  \log p_t(X_t)$ in (\ref{eq_our_cond_score_main_tex}) with $s_{\theta}(X_t, t, \varnothing)$, we get Diffusion Model with Double Guidance (DMDG):
\begin{align}
\label{eq_double_guidance_approx}
    & \nabla \log p_t(X_t|C_1, C_2) \notag \\
    & \approx s_{\theta}(X_t, t, \varnothing) - \lambda_1 \nabla \|C_1-\widehat{f}_1(X_{0|t}) \|_2^2 \notag \\
    & \quad - \lambda_2 \nabla \|C_2-\widehat{f}_2(X_{0|t,C_1}) \|_2^2.
\end{align}
Compared to DPS, our DMDG introduces an additional guidance term related to $C_2$, $- \lambda_2 \nabla \| C_2 - \widehat{f}_2(X_{0 | t, C_1}) \|_2^2$. This term enables explicit control over $C_2$, while implicitly preserving the correlation between $C_1$ and $C_2$ through $\widehat{f}_2(X_{0 | t, C_1})$. In addition, DMDG does not require joint samples $(X_0, C_1, C_2)$ because the regressors (or classifiers) $\widehat{f}_1$ and $\widehat{f}_2$ can be trained on distinct datasets separately. Thus, DMDG circumvents the challenges posed by block-wise missing structure in aggregated datasets. Substituting $\widehat{f}_2(X_{0|t,C_1})$ in (\ref{eq_double_guidance_approx}) with $\widehat{f}_2(X_{0|t})$ results in DMIDG. The detailed derivation of (\ref{eq_our_cond_score_main_tex}) is presented in Appendix \ref{sec_app_proof_and_derivation}, which does not violate $C_1 \not \perp \!\!\! \perp C_2 | X_t$.

\subsection{Diffusion Model with Hybrid Guidance (DMHG)}
Similarly to CFG, we can replace $ \nabla \log p(C_1|X_t)$ in (\ref{eq_our_cond_score_main_tex}) by $\lambda_1 (\nabla \log p_t(X_t|C_1) - \nabla \log p_t(X_t))$. Following the first equality in~(\ref{eq_our_cond_score_main_tex}) and the Gaussian modeling in (\ref{relationship_nerual_estimator}), we obtain:
\begin{align}
\label{eq_hybrid_guidance}
    & \nabla \log p_t(X_t | C_1, C_2)  \notag \\
    & \approx (1-\lambda_1)s_{\theta}(X_t, t, \varnothing) + \lambda_1 s_{\theta}(X_t, t, C_1) \notag \\
    & \ \ \ \  - \lambda_2 \nabla \|C_2-\widehat{f}_2(X_{0|t,C_1}) \|_2^2.
\end{align}
We term this approach the Diffusion Model with Hybrid Guidance (DMHG), as it incorporates both CG and CFG. Similarly to CFG, DMHG exhibits improved fidelity to the target conditional distribution
$P_{X_0 | C_1, C_2}$ when condition $C_1$ is highly complex. Similarly to DMIDG, replacing $\widehat{f}_2(X_{0|t,C_1})$ in (\ref{eq_hybrid_guidance}) with $\widehat{f}_2(X_{0|t})$ results in Diffusion Model with Independent Hybrid Guidance (DMIHG). The complete sampling algorithm for both DMDG and DMHG is provided in Algorithm \ref{algo_dmdg_and_dmhg}. Extensions of DMDG and DMHG to settings involving more than two aggregated datasets and multiple conditions are presented in Appendix~\ref{sec_dmtg}.

\section{Theoretical Analysis}
\label{sec_thm}
DMDG proceeds by first approximating $p(C_1 | X_t)$ and $p(C_2 | X_t, C_1)$ in~(\ref{eq_our_cond_score_main_tex}) with $p(C_1 | \mathbb{E}[X_0 | X_t])$ and $p(C_2 | \mathbb{E}[X_0 | X_t, C_1])$, respectively. These densities are then further approximated by $p(C_1 | X_{0 | t})$ and $p(C_2 | X_{0 | t, C_1})$, as defined in~(\ref{relationship_nerual_estimator}). Thus, the quantities of $|p(C_1|X_t) - p(C_1|X_{0|t}) |^2$ and $|p(C_2|X_t,C_1) - p(C_2|X_{0|t,C_1})|^2$ serve as measures of the effectiveness of the proposed methods. Theorem \ref{thm_square_of_densities} characterizes the key factors that contribute to these discrepancies. All proofs are provided in Appendix \ref{sec_app_proof}. 
\begin{theorem}
\label{thm_square_of_densities}
    Assume that $p(C_1|X_t)$ and $p(C_2|X_t,C_1)$ denote the densities of $C_1$ given $X_t$, and of $C_2$ given $(X_t,C_1)$, respectively. Further assume that $p(C_1|X_{0|t})$ and $p(C_2|X_{0|t,C_1})$ are Gaussian densities defined in (\ref{relationship_nerual_estimator}). Assume that $\widehat{f}_1$ and $\widehat{f}_2$ in (\ref{relationship_nerual_estimator}) are $L$-layer ReLU networks. Let $M$ denote the maximum $\ell_2$-norm of their weight matrices and let $U_1$ and $U_2$ be positive finite constants. Then, we have:
    \begin{align}
        &  | p(C_1|X_t)-p(C_1|X_{0|t}) |^2   \label{eq_thm1_1st_ineq}  \\
        &   \leq \dfrac{k}{\sqrt{2 \pi} \sigma_{1}} \exp \left( - \dfrac{1}{2 \sigma_{1}^2} \right)  \Bigl\{ (2 \max_{z \in \mathbb{R}^d} \| \nabla_{z} f_1(z) \|_2^2 + M^L) \notag \\
        & \quad \cdot \mathbb{E}_{X_0 \sim P_{X_0|X_t}} \| X_0 - \mathbb{E} \left[ X_0|X_t \right] \|_2^2   \notag \\
        &  \quad  + M^L \| \mathbb{E}\left[ X_0|X_t \right] -X_{0|t} \|_2^2 \notag \\
        & \quad + \mathbb{E}_{X_0 \sim P_{X_0|X_t}} \| f_1(X_0)-\widehat{f}_1(X_0) \|_2^2 \Bigr\} \notag \\
        & \quad + U_1 (\sigma_1^2-1/2\lambda_1)^2, \notag
    \end{align}
and
\begin{alignat}{2}
        & | p(C_2|X_t,C_1) - p(C_2|X_{0|t,C_1}) |^2 \label{eq_thm1_2nd_ineq} \\
        & \leq \dfrac{p-k}{\sqrt{2 \pi} \sigma_{2}} \exp \Bigl( \dfrac{-1}{2 \sigma_{2}^2} \Bigr) \Bigl\{ (2 \max_{z \in \mathbb{R}^d} \| \nabla_{z} f_2(z) \|_2^2 + M^L)  \hspace*{0.6em}\text{\bf\small \textit{a) b)}}\notag \\
        & \ \ \ \ \ \cdot \mathbb{E}_{X_0 \sim P_{X_0|X_t,C_1}} \| X_0 - \mathbb{E} \left[ X_0|X_t,C_1 \right] \|_2^2  \hspace*{1.74cm}\text{\bf\small \textit{c)}} \notag \\
        & \ \ \ \ + M^L \| \mathbb{E}\left[ X_0|X_t,C_1 \right] -X_{0|t,C_1} \|_2^2 \hspace*{2.65cm}\text{\bf\small \textit{d)}} \notag \\
        & \ \ \ \ + \mathbb{E}_{X_0 \sim P_{X_0|X_t,C_1}} \| f_2(X_0)-\widehat{f}_2(X_0) \|_2^2 \Bigr\} \hspace*{1.85cm}\text{\bf\small \textit{e)}} \notag \\
        & \ \ \ \ + U_2 (\sigma^2_2-1/2\lambda_2)^2 \hspace*{4.8cm}\text{\bf\small \textit{f)}} \notag
    \end{alignat}
\end{theorem}

\begin{remark}
    we consider~(\ref{eq_thm1_2nd_ineq}) as an illustrative example. The discrepancy between the true density $p(C_2 | X_t, C_1)$ and its practical approximation $p(C_2 | X_{0 | t, C_1})$ defined in~(\ref{relationship_nerual_estimator}) is determined by the following factors: \textbf{a)} the magnitude of the variance $\sigma^2_2$; \textbf{b)} the maximum norms of the gradients $\nabla_{z} f_2(z)$ and $ \nabla_{z} \widehat{f}_2(z) $; \textbf{c)} the trace of $\operatorname{Cov}(X_0|X_t,C_1)$; \textbf{d)} the discrepancy between $\mathbb{E} \left[ X_0|X_t,C_1 \right]$ and $X_{0|t,C_1}$; \textbf{e)} the averaged $\ell_2$-norm between $f_2$ and $\widehat{f}_2$, given $X_t$ and $C_1$; \textbf{f)} the discrepancy between the true variance $\sigma^2_2$ and the variance $1/2\lambda_2$ assigned to $p(C_2|X_{0|t,C_1})$. 
    
    Item \textbf{a)} indicates that when $ \sigma^2_2 $ is large or close to zero, a large part of the error diminishes. The $\ell_2$-norms in \textbf{b)} and \textbf{c)} are generally bounded, even when they remain unspecified \cite{chung_diffusion_inverse}. In addition, \textbf{c)} vanishes to zero as $t \to 0$. The magnitude of \textbf{d)} is influenced by the accuracy of EDM's neural network, as implied by the loss function \eqref{eq_edm_loss}. \textbf{e)} is related to the conditional prediction error of $\widehat{f}_2$, given $X_t$ and $C_1$. Finally, \textbf{f)} reveals that an inappropriate selection of the guidance scale $\lambda_2$ can introduce additional errors, and the appropriate scale is $\lambda_2 \sim \mathcal{O}(\sigma_2^{-2})$, which is confirmed by our extra simulations in Appendix \ref{sec_app_gaussian_mix}.
\end{remark}

\section{Molecule Generation Tasks}
\label{sec_mol}

\subsection{The aggregated datasets}
We aggregate two datasets for molecular generation: GEOM-DRUG and ZINC250k. GEOM-DRUG 
contains more than 210,000 drug-like molecular conformers annotated with their energetic properties \cite{Axelrod_2022_geomdataset}. We focus on lowest energy, ensemble energy, and ensemble entropy, which characterize molecular stability and flexibility required for receptor binding. An ideal drug candidate typically exhibits low values of the first two and moderately low values of the third \cite{Fogolari_2018_energy_lower_better}. The latter dataset, ZINC250k consists of approximately 250,000 molecules annotated with drug-likeness properties \cite{Sterling_2015_zincdataset}. We selected logP, QED, and SAS as representative properties, as promising drug candidates are generally expected to meet desirable thresholds of these metrics \cite{Ertl_2009_sas_is_important}.
We aggregate GEOM-DRUG and ZINC250k to train the diffusion models, and separately train $\widehat{f}_1$ and $\widehat{f}_2$ using their respective datasets.

\subsection{De novo drug design using DMDG}
\label{sec_design_mole_from_0}
De novo drug design refers to the generation of entirely new molecular structures. 
The aim is to generate molecules that exhibit both energetic stability and high drug-likeness.  We use the proposed diffusion models with two datasets; 
we set $D^{(1)}$ as GEOM-DRUG, with $C_1  = (\text{lowest energy}, \text{ensemble energy}, \text{ensemble entropy})$, and $D^{(2)}$ as ZINC250k, with $C_2 = (\text{logP}, \text{QED}, \text{SAS})$. We consider two tasks with the following conditions:
\begin{align}
\label{eq_molecular_task_12}
    \text{Task 1:} &\text{ Lowest energy, Ensemble energy} \in [q_{0.1}, q_{0.4}], \notag \\
    \textcolor{white}{\text{Task 1:}} & \text{ Ensemble entropy} \in [q_{0.25},q_{0.5}]; \notag \\
    \text{Task 2:}& \text{ logP} \in [0,4], \text{QED} \in [0.7,1], \text{SAS} \in [0,5],  
\end{align}
where each $q_\alpha$ in Task 1 denotes the $\alpha$-quantiles of the corresponding property. 

We compare DMDG and DMHG with a diverse set of baselines, including the CG-based DMIDG \cite{ye_2024_unified_guidance} and the CFG-based DMIHG.  We also consider methods based on compositional score functions, such as Composition \cite{Liu_2022_compositional_diff}. In addition, we include CTRL \cite{zhao_2025_adding_cond_by_rl}, which leverages RL and ControlNet to fine-tune a pretrained conditional diffusion model. We further evaluate four imputation-based approaches, which first recover missing conditions in the aggregated datasets and then train a CDM: regressor imputation, GAIN \cite{Yoon_2018_gain}, Forest Diffusion \cite{Jolicoeur_2024_forestdiffusion}, and KNN imputation. All diffusion models operate in the latent space learned by COATI \cite{Kaufman_2024_COATI}, a powerful variational autoencoder (VAE).  For each method, we generate 5{,}000 molecules, with their properties guided toward the midpoints of the specified intervals. A generation is considered successful if all the six molecular properties \textit{simultaneously} satisfy the constraints of Tasks~1 and~2.  In addition, we evaluate the $W_2$ distances between the generated molecules and the target conditional distributions. Unless otherwise specified, all diffusion models use 18 sampling steps, following \citet{karras_edm}. Further details of the implementation and compared methods are provided in Appendix~\ref{sec_compared_methods}.

\begin{figure*}[ht!]
  \centering
  \includegraphics[width=0.6\textwidth]{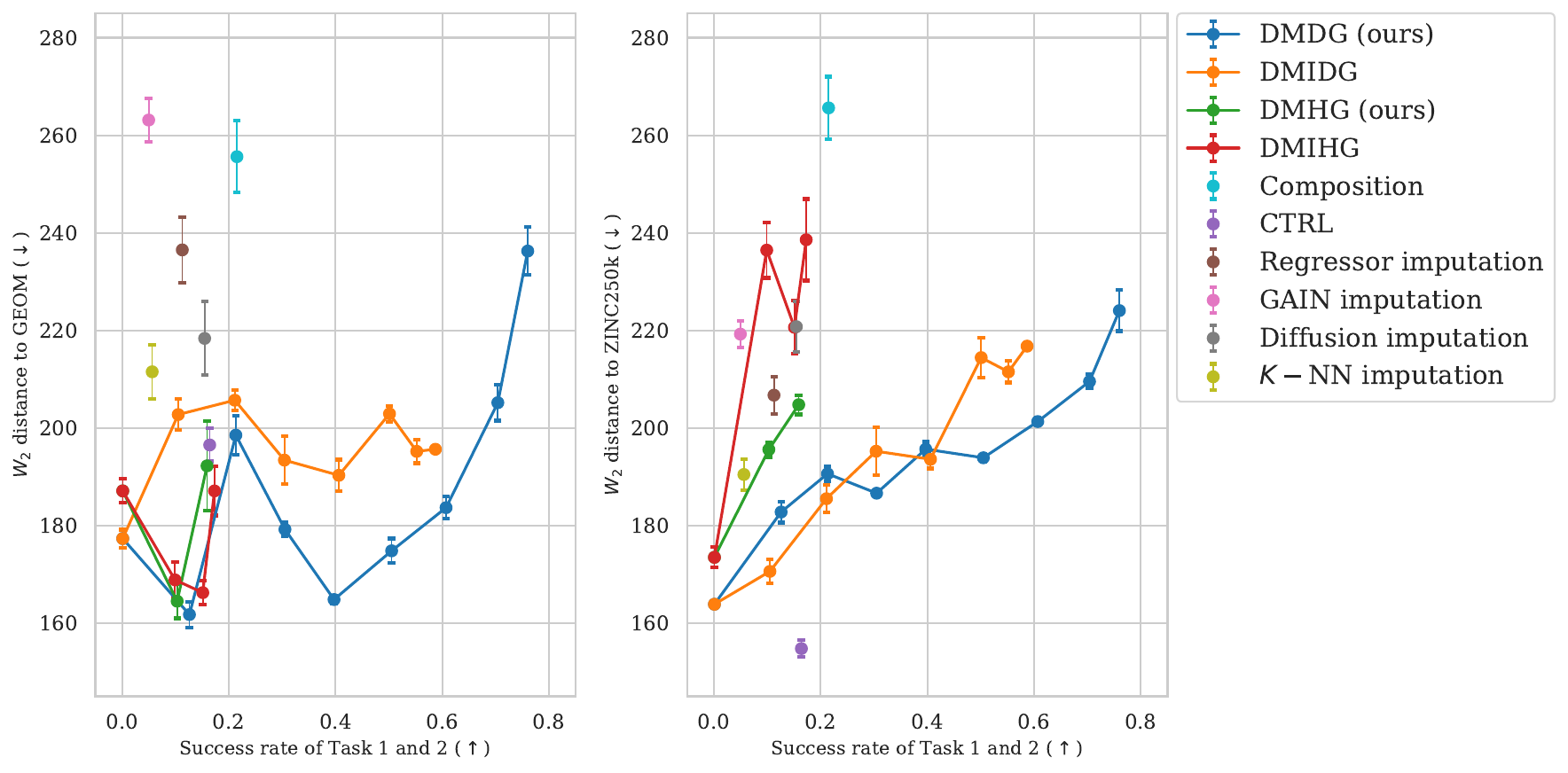}
  \caption{Success rate of Tasks 1 \& 2 (\ref{eq_molecular_task_12}) and $W_2$ distances to GEOM-DRUG and ZINC250k. The results of DMDG, DMIDG, DMHG, and DMIHG are shown as curves, since they involve tunable parameters $\lambda_1$ and $\lambda_2$. Error bars represent one standard deviation across 5000 generated molecules.}
  \label{fig_molecular_task_12}
\end{figure*}

The results are reported in Figure \ref{fig_molecular_task_12}. CG-based methods, DMDG and DMIDG, achieve relatively high success rates. DMDG outperforms DMIDG by a substantial margin (76.0\% vs. 58.7\%), highlighting the importance of accounting for the dependence
between $C_1$ and $C_2$ given $X_t$.
At the same success rate, DMDG also attains a significantly smaller $W_2$ distance from the GEOM-DRUG dataset. In contrast, CFG-based methods exhibit limited improvement in task success, with all success rates remaining below 25\%, consistent with the observations reported by \citet{Kaufman_2024_coatildm}. Compositional and imputation-based methods similarly demonstrate limited effectiveness, with success rates also generally below 25\%. CTRL, the RL-based approach fine-tuned on ZINC250k, achieves a relatively low $W_2$ distance from ZINC250k; however, its success rate remains below 20\%.

\begin{table*}[ht!]
\centering
\caption{Comparison of success rates of different models under different Calls.}
\begin{tabular}{lccccc}
\toprule
\textbf{Method} & \textbf{Call=25} & \textbf{Call=100} & \textbf{Call=500} & \textbf{Call=1000} & \textbf{Call=50000}  \\
\midrule
DMDG (ours) &  37.50 \% &  57.12 \% &  75.62 \% & 79.87 \% &   94.37 \%  \\
DMIDG \cite{ye_2024_unified_guidance} &  36.12 \% &  44.25 \% &  65.37\% & 74.87\% &  94.37 \%  \\
DPS \cite{chung_diffusion_inverse} &  32.75 \% &  43.75 \% &  65.37 \% & 74.25 \% &  94.25 \%  \\
COATI-LDM  \cite{Kaufman_2024_coatildm} & 0.62 \% & 4.91 \% & 45.27 \% &84.44 \%& 96.2 \%  \\
DESMILES \cite{Maragakis_2020_DESMILES} & N/A& N/A&N/A &N/A & 77.2 \%  \\
QMO \cite{Hoffman_2022_QMO} &N/A &N/A &N/A &N/A &  92.4 \%  \\
\bottomrule
\end{tabular}
\label{tab_method_call_results}
\end{table*}

\subsection{Nearby sampling using DMDG.} 
\label{sec_nearby_sampling}
In drug discovery, it is common to start from a set of promising source molecules and seek new structures that satisfy all desired properties while remaining similar to the original structure. This paradigm is commonly referred to as nearby sampling.  
In the context of diffusion models, nearby sampling can be accomplished by partial diffusion \cite{Kaufman_2024_coatildm}, which involves adding limited noise to a source molecule through (\ref{eq_forward_process}), followed by a reverse process guided by the desired conditions. 

Following the previous work \cite{jin_2018_g2g,Kaufman_2024_coatildm,ikeda_2025_a2a_transports}, we consider source molecules with initial QED scores in $[0.6, 0.8]$. Our goal is to generate molecules that achieve QED $\geq 0.9$ after partial diffusion, while maintaining a Tanimoto similarity of at least $0.4$ to their source molecules. 
For each source molecule, we perform multiple \emph{Calls}, where each Call corresponds to a single generation attempt.
A trial is considered successful if at least one molecule in the batch satisfies the two targets. 

Unlike the setting in Section~\ref{sec_design_mole_from_0}, we set $D^{(1)}$ to ZINC250k and $D^{(2)}$ to GEOM-DRUG. The condition $C_1$ corresponds to QED, while $C_2$ denotes the Tanimoto similarity to the source molecule. To calculate the gradient, a regressor $\widehat{f}_2$ is trained to estimate the Tanimoto similarity between two molecules. Details can be found in Appendix \ref{sec_app_tanimoto_regressor}. In the reverse process, the target values for QED and Tanimoto similarity are set to $1.0$ and $0.7$, respectively.

Given the poor performance of CFG-based methods shown in Figure \ref{fig_molecular_task_12}, we restrict our experiment to CG-based models, namely DMDG (ours), DMIDG, and DPS. In addition, we compare several recent state-of-the-art models, including COATI-LDM, a large molecular generative model pre-trained on around 250 million molecules \cite{Kaufman_2024_coatildm}; DESMILES, an RNN-based model \cite{Maragakis_2020_DESMILES}; and QMO, a seq-to-seq model \cite{Hoffman_2022_QMO}.

We randomly select 800 source molecules from  ZINC250k and compare the success rates of different methods under varying number of Calls (batch sizes). As shown in Table \ref{tab_method_call_results}, DMDG, DMIDG, and DPS achieve higher success rates than the pre-trained COATI-LDM in the low-Call regime, particularly when the number of Calls is below 500. It is worth noting that DMDG, DMIDG, and DPS use a fixed number of 18 sampling steps, whereas COATI-LDM employs 1000 sampling steps. Among the CG-based methods, DMDG consistently achieves the highest success rates, demonstrating the effectiveness of the proposed approach.

Overall, our proposed DMDG demonstrates strong controllability over conditions on aggregated datasets while reducing computational cost. The results show that DMDG has the potential to benefit the drug discovery pipeline. Further experiments are provided in Appendix \ref{sec_additional_exps}. Implementation details can be found in Appendix \ref{sec_experimental_details}.   Representative generated molecules are presented in Appendix \ref{sec_selected_mols}.

\begin{figure*}[ht!]
    \centering
    \begin{minipage}[b]{0.48\linewidth}
        \centering
        \includegraphics[width=\linewidth]{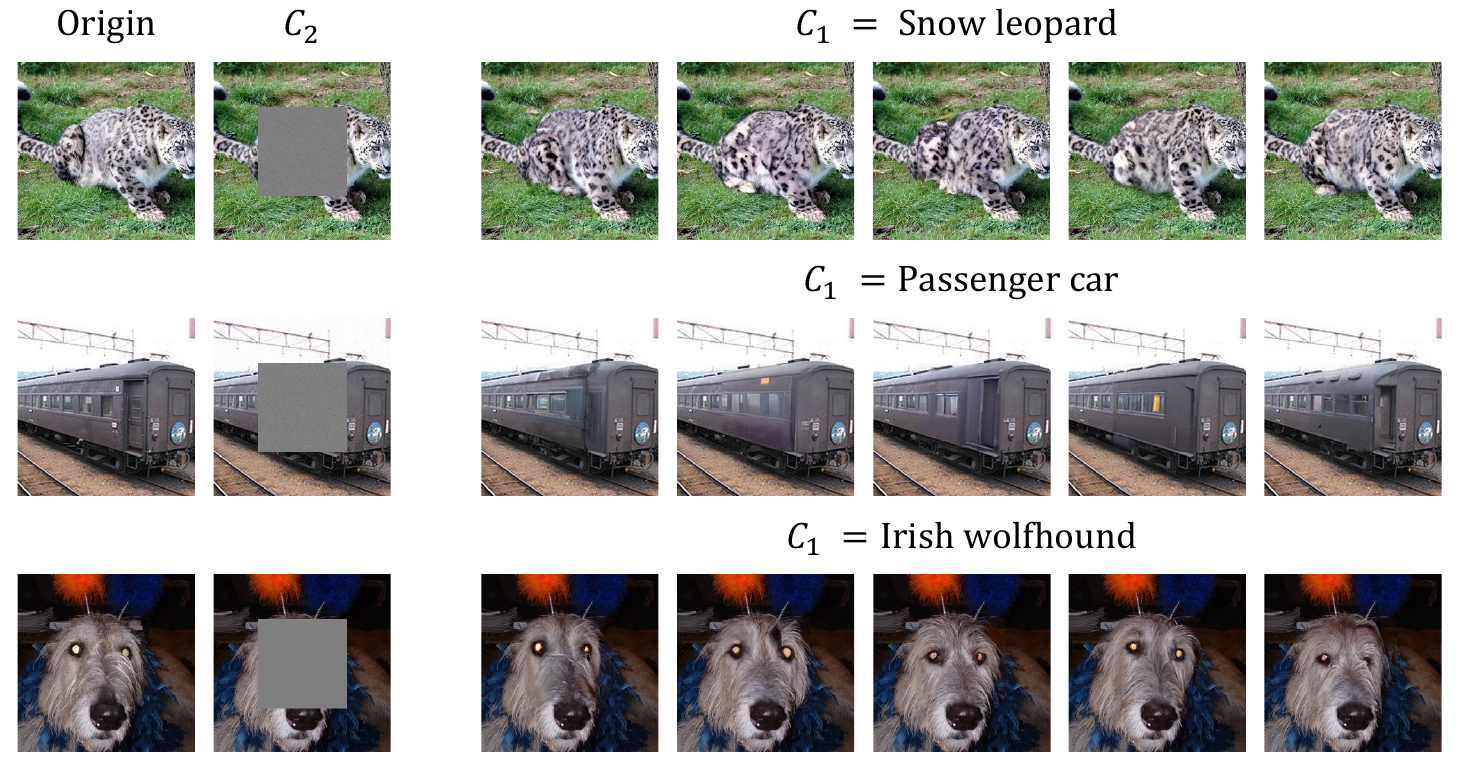}

    \end{minipage}
    \hfill
    \begin{minipage}[b]{0.48\linewidth}
        \centering
        \includegraphics[width=\linewidth]{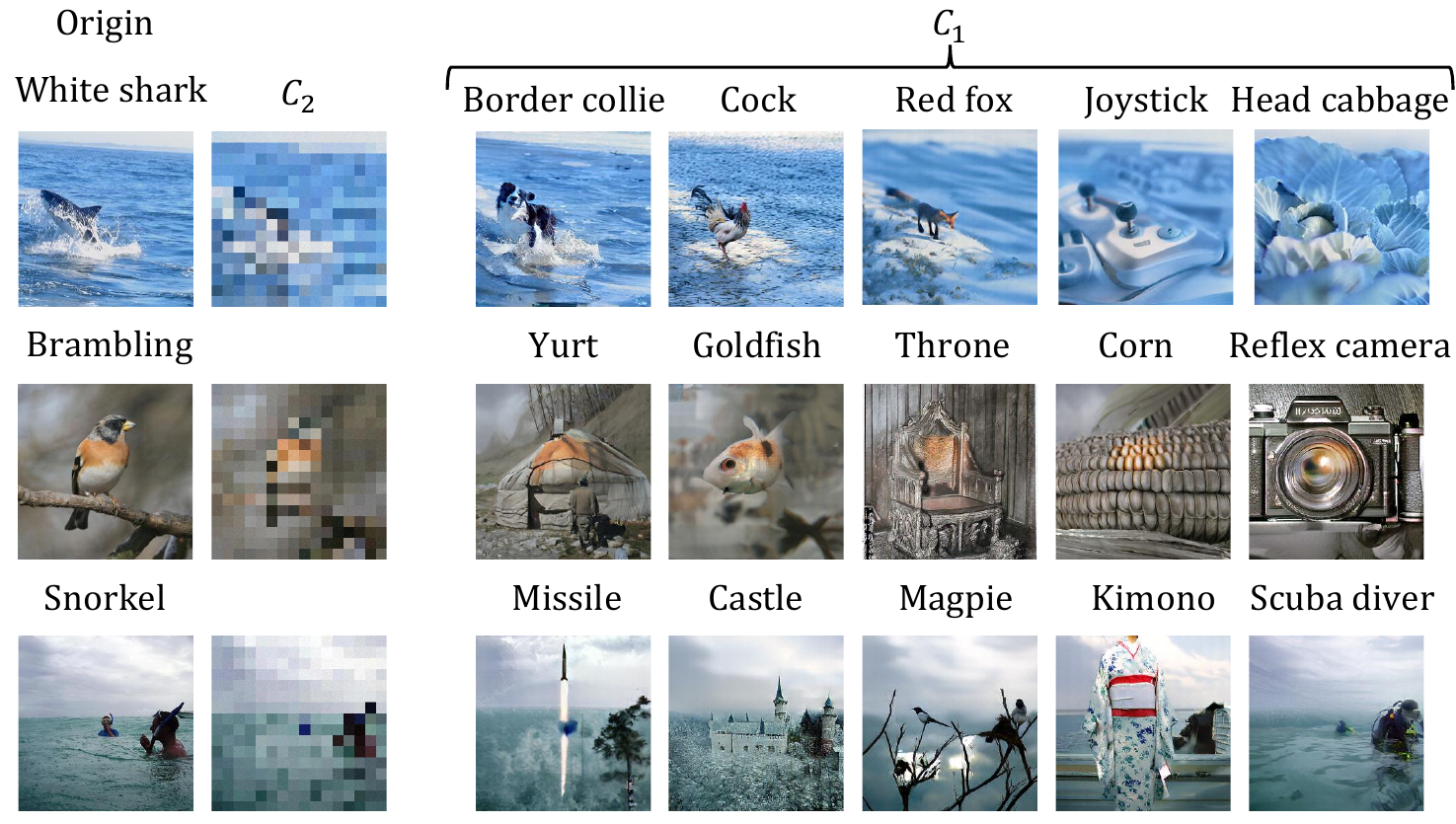}

    \end{minipage}
    \caption{Left panel: the results of image inpainting via DMHG. The guidance scales of inpainting are $\lambda_1 = 2.5$ and $\lambda_2 = 100$. Right panel: the results of object editing via DMHG. The guidance scales of object editing are $\lambda_1 = 4.5$ and $\lambda_2 = 200$.}
    \label{fig_inpaint_obj_edit}
\end{figure*}

\section{Image Generation Tasks}
\label{sec_image_gen}
To demonstrate the applicability of our approach across different data modalities,
we additionally conduct image generation experiments. As discussed in Section~\ref{sec_problem_set}, when $X_0$ is an image, the function $f_1$ can be interpreted as an object-mapping function, while $f_2$ corresponds to an image transformation, such as masking or down-sampling. Accordingly, $C_1$ represents the category of the object in $X_0$, and $C_2$ denotes the transformed $X_0$. Depending on the choice of $f_2$, we consider two applications: image inpainting when $f_2$ is a mask, and object editing when $f_2$ corresponds to down-sampling.
While these image generation tasks do not involve aggregated datasets,
they serve to validate the general applicability of the proposed
double guidance framework, which can be directly applied in this setting.

We uniformly adopt the JIT-H model \cite{li_2025_jit} pre-trained on ImageNet-256, with the number of sampling steps fixed to 50 throughout all experiments in this section. Given the large scale of ImageNet and the strong performance of the JIT model,
we do not aggregate additional datasets.
As discussed in Section~\ref{sec_cfg}, CFG-based methods are more suitable for image generation tasks;
therefore, we restrict our comparison to our proposed DMHG, DMIHG, as well as DPS. The object-mapping function $f_1$ is approximated by $\widehat{f}_1$, a fine-tuned ResNet-18, while the transformation function $f_2$, masking or down-sampling, is assumed to be known. In addition, because JIT is a generative model based on flow matching \cite{lipman_2023_fm}, we provide a detailed instruction in Appendix~\ref{sec_app_fm_to_guidance} on how CG and CFG can be applied within the FM framework. A selection of generated images are shown in Figure \ref{fig_inpaint_obj_edit} and Figures \ref{fig_inpaint_edit} and \ref{fig_img_more_examples} in Appendix.

\subsection{Better image inpainting}
\label{sec_inpaint}
Image inpainting is a task with broad practical applications \cite{chung_diffusion_inverse, Lugmayr_2022_repaint}. When $f_2$ corresponds to a mask, the inpainting task aims to recover the masked regions of $X_0$ given $C_2$. However, when the category $C_1$ of $X_0$ is available (which often occurs when the mask does not fully occlude the object), $C_1$ can be incorporated into the inpainting to further improve performance. We set $f_2$ to a $128 \times 128$ mask that occludes the center of the image. From each ImageNet class, two images are randomly sampled, yielding 2{,}000 images for this task. 

The results of image inpainting are shown in the top panel of Figure \ref{fig_inpaint_edit_lpips} in Appendix. Both DMHG and DMIHG outperform DPS in all five evaluation metrics. While DMHG and DMIHG exhibit comparable control over conditions $C_1$ and $C_2$, DMHG achieves a lower LPIPS \cite{Zhang_2018_lpips}, indicating that the images inpainted by DMHG are closer to the ground-truth images
Detailed discussion can be found in Appendix \ref{sec_app_inpaint}. The left panel of Figure~\ref{fig_inpaint_obj_edit} presents several examples of image inpainting.
As can be observed, DMHG produces high-quality images
with diverse and realistic content in the inpainted regions.

\subsection{Object editing}
\label{sec_obj_edit}
We next consider a super-resolution inverse problem with semantic modification, also referred to as object editing. Here, $f_2$ is a down-sampling operation ($\times 16$). Given a low-resolution observation of an image containing a source object, the task is to generate a high-resolution image of a different target object using DMHG and DMIHG. In this experiment, 500 images are randomly selected and edited to a different object.

The object editing results are shown in the bottom panel of Figure~\ref{fig_inpaint_edit_lpips} in Appendix. While DMHG and DMIHG achieve comparable LPIPS and alignment to $C_2$, DMHG exhibits significantly stronger control over $C_1$ than DMIHG. Additional discussions and qualitative results are provided in Appendix~\ref{sec_app_edit}. In addition, when $f_2$ is a mask, DMHG can also perform object editing. However, this requires manual placement of the mask in every image, making quantitative evaluation impossible. Therefore, we only present some representative examples of inpainting-based object editing in Appendix~\ref{sec_app_obj_edit_and_inpaint}. The right panel of Figure~\ref{fig_inpaint_obj_edit} shows examples of image editing,
demonstrating that DMHG can generate high-resolution images with multiple different objects given a low-resolution image.

\section{Conclusion}
\label{sec_conclusion}
In this paper, we propose a double guidance framework and two specific instantiations, DMDG and DMHG, to enable joint conditional generation on aggregated datasets with block-wise missing. We established a more detailed theoretical analysis of guided diffusion. The proposed DMDG achieves superior joint-condition controllability in two molecular generation tasks, and DMHG is able to generate high-quality images conditioned on both category and transformed image. Combining our methods with more advanced techniques of guided diffusion, self-adaptive guidance scales, and deploying them on larger and more aggregated datasets constitute our future directions.


\section*{Impact Statement}

This work aims to advance the methodology of conditional generative modeling
by enabling joint conditional generation on aggregated datasets
with block-wise missing conditions.
Our proposed methods have potential applications in domains such as
drug discovery, where they may facilitate more efficient exploration
of molecular design spaces under multiple constraints.
In addition, the image generation experiments demonstrate that the proposed
approach can generalize to other data modalities, such as image editing,
and may be extended to tasks including video editing in the future.

This work focuses on methodological contributions to generative modeling
and does not involve human subjects, personal data, or direct real-world deployment.
We do not foresee immediate negative societal consequences arising from this work.

\bibliography{icml_2026}
\bibliographystyle{icml2026}

\newpage
\appendix
\onecolumn
\section{Derivation and Algorithm of DMDG and DMHG}
\label{sec_app_proof_and_derivation}
\begin{figure}[ht!]
    \centering
    \begin{subfigure}[t]{0.4\textwidth}
        \centering
        \includegraphics[width=\linewidth]{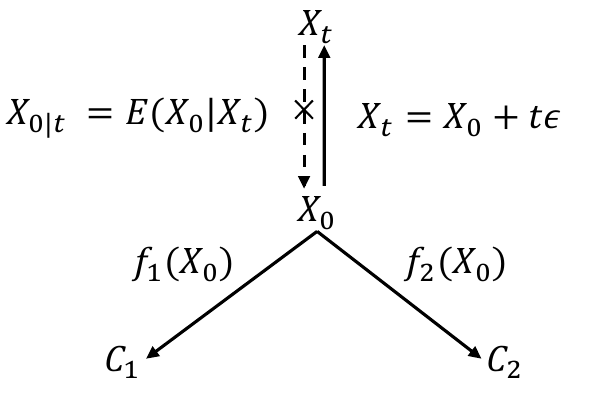}
    \end{subfigure}
    \begin{subfigure}[t]{0.36\textwidth}
        \centering
        \includegraphics[width=\linewidth]{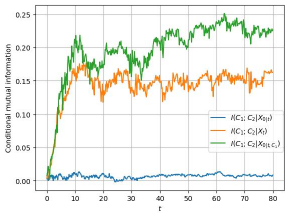}
    \end{subfigure}
    \caption{Left panel: Conditional independence relationships among $X_0, X_t,C_1$ and $C_2$: $X_t, C_1,C_2$ are mutually independent given $X_0$, and $C_1 \perp \! \! \! \perp C_2 |X_{t}$ and $ C_1 \perp \! \! \! \perp C_2 |X_{0|t}$ are not true. In addition, DMDG and DMHG do not assume either $C_1 \perp\!\!\!\perp C_2$ or $C_1 \not\!\perp\!\!\!\perp C_2$, as our proposed methods are applicable to both cases. Right panel: Under the setting II in Appendix \ref{sec_app_gaussian_mix}, the behavior of conditional mutual informations (CMI) over the scale of $t$ in forward process (\ref{eq_forward_process}). }
    \label{fig_CI_relationship}
\end{figure}

First, we show how to approximate the conditional score $\nabla \log p_t(X_t|C_1,C_2)$ in (\ref{eq_our_cond_score_main_tex}):
\begin{align}
\label{eq_our_cond_score_appendix}
    &\nabla \log p_t(X_t|C_1, C_2) \notag \\
    &= \nabla \log  p_t(X_t) + \nabla \log  p(C_1,C_2|X_t)  \notag \\
    &= \nabla \log  p_t(X_t) + \nabla \log  p(C_2|X_t,C_1) + \nabla \log  p(C_1|X_t)  \notag \\
    &= \nabla \log  p_t(X_t)  + \nabla \log \int  p(C_2,X_0|X_t,C_1) dX_0  + \nabla \log \int  p(C_1,X_0|X_t) dX_0   \notag \\
    &= \nabla \log  p_t(X_t)  + \nabla \log \int  p(C_2|X_0,X_t,C_1)p(X_0|X_t,C_1) dX_0 + \nabla \log \int  p(C_1|X_0,X_t)p(X_0|X_t) dX_0   \notag \\
    &\overset{(*)}{=} \nabla \log  p_t(X_t)  + \nabla \log \int  p(C_2|X_0)p(X_0|X_t,C_1) dX_0 + \nabla \log \int  p(C_1|X_0)p(X_0|X_t) dX_0   \notag \\
    &\overset{(**)}{\approx} \nabla \log p_t(X_t) + \nabla \log p(C_2|\mathbb{E} \left[ X_0|X_t,C_1 \right]) + \nabla \log p(C_1|\mathbb{E} \left[ X_0|X_t \right]).
\end{align}
In $(*)$, we use the property that $C_2 \perp \!\!\! \perp (X_t,C_1) | X_0$, as shown in Figure \ref{fig_CI_relationship}. For $(**)$, following \cite{chung_diffusion_inverse}, we approximate $\int p(C_1|X_0)p(X_0|X_t)dX_0$ and $ \int  p(C_2|X_0)p(X_0|X_t,C_1) dX_0$ by:
\begin{align}
    & \int p(C_1|X_0)p(X_0|X_t)dX_0 = \mathbb{E}_{X_0 \sim P_{X_0|X_t}}\left[ p(C_1|X_0) \right] \approx p(C_1|\mathbb{E} \left[ X_0|X_t \right]), \notag \\
    & \int p(C_2|X_0)p(X_0|X_t,C_1) dX_0 = \mathbb{E}_{X_0 \sim P_{X_0|X_t,C_1}}\left[ p(C_2|X_0) \right] \approx p(C_2| \mathbb{E} \left[ X_0|X_t,C_1 \right] ). \notag
\end{align}

The algorithm of DMDG and DMHG can be found in Algorithm \ref{algo_dmdg_and_dmhg}.

\begin{algorithm*}[ht!]
\caption{Conditional sampling by DMDG or DMHG}
\label{algo_dmdg_and_dmhg}
\textbf{Input}: Target conditions $C_1$, $C_2$, a trained score network $nn_{\theta}$, regressors $\widehat{f}_1, \widehat{f}_2$, guidance scale $\lambda_1, \lambda_2$, a time schedule $t_0, t_2, \hdots, t_N$ with $t_0=t_{\min}, t_N=t_{\max}$, sampling method: DMDG or DMHG.\\
\textbf{Output}: A sample approximately follows $P_{X_0|C_1,C_2}$.
\begin{algorithmic}[1] 
\STATE Draw $X_{t_{N}} \sim N(0,t_N^2 I_d)$
\FOR{$i$ in $N-1, N-2, \hdots, 0$}
\STATE Let $X_{0|t_{i+1}}=nn_\theta(X_{t_{i+1}},t_{i+1},\varnothing)$, $X_{0|t_{i+1},C_1}=nn_\theta(X_{t_{i+1}},t_{i+1},C_1)$
\STATE Let $s_{i+1,\varnothing}=\dfrac{X_{0|t_{i+1}}-X_{t_{i+1}}}{t_{i+1}^2}$, $s_{i+1,C_1}=\dfrac{X_{0|t_{i+1},C_1}-X_{t_{i+1}}}{t_{i+1}^2}$
\IF{sampling method = DMDG}
\STATE Let $d_{i+1}=s_{i+1,\varnothing}-\lambda_1 \nabla_{X_{t_{i+1}}} \| C_1- \widehat{f}_1(X_{0|t_{i+1}}) \|_2^2 - \lambda_2 \nabla_{X_{t_{i+1}}} \| C_2- \widehat{f}_2(X_{0|t_{i+1},C_1}) \|_2^2 $
\ELSIF{sampling method = DMHG}
\STATE Let $d_{i+1}=(1-\lambda_1)s_{i+1,\varnothing}+\lambda_1 s_{i+1,C_1}  - \lambda_2 \nabla_{X_{t_{i+1}}} \| C_2- \widehat{f}_2(X_{0|t_{i+1},C_1}) \|_2^2 $
\ENDIF
\STATE $X_{t_{i}} = X_{t_{i+1}}-(t_{i}-t_{i+1}) \cdot t_{i+1} \cdot d_{i+1}$
\ENDFOR
\STATE \textbf{return} $X_{t_{0}}$
\end{algorithmic}
\end{algorithm*}

\section{Extra Simulation: A Gaussian Mixture Example}
\label{sec_app_gaussian_mix}
To enable a more analytic evaluation of the generation performance,
we conduct simulations on a Gaussian mixture model. The target variable is $X_0 \in \mathbb{R}^{5}$, 
$X_0 \sim \sum_{k=1}^{K} w_k N(m_k,\Sigma_k)$ ($K = 10$)  
and the function mapping is a linear transformation:
\begin{equation*}
    Y:=(Y_1,Y_2,Y_3,Y_4,Y_5)^\top  = AX_0+\sigma \eta,  \eta \sim N(0,I_5),
\end{equation*}
where $A \in \mathbb{R}^{5 \times 5}$ is a fixed matrix and $\sigma>0$ is the standard deviation. Given $Y=y$, the posterior distribution of $X_0$ can be analytically computed:
\begin{equation} \label{eq:exact_poster_GM}
{\textstyle 
    p_{X_0|Y=y} \propto \sum_{k=1}^{K} \widetilde{w}_k N(\widetilde{m}_k,\widetilde{\Sigma}_k), }
\end{equation}
where $\widetilde{\Sigma}_k = ( A^\top A/\sigma^2+ \Sigma_k^{-1}  )^{-1}$, $\widetilde{m}_k = \widetilde{\Sigma}_k (A^\top y/\sigma^2+ \Sigma_k^{-1} m_k )$,
and 
$
    \widetilde{w}_k =  w_k \cdot \exp [ 0.5(\widetilde{m}_k^\top  \widetilde{\Sigma}_k^{-1} \widetilde{m}_k    -   m_k^\top  \Sigma_k^{-1} m_k) ] / \sqrt{|\Sigma_k|}$ \cite{Hagemann_2022_gaussian_simulation}.
We set $C_1 =(Y_1,Y_2)^\top $, $C_2 =(Y_3,Y_4,Y_5)^\top $, and each dataset $D^{(1)}$ and $D^{(2)}$ contains 10000 samples. 
With \eqref{eq:exact_poster_GM}, we can accurately evaluate the distance between the generated samples and the true conditional distribution.

The baselines are the same as those mentioned in Section~\ref{sec_design_mole_from_0}. We consider the following two settings:

\noindent \textbf{Setting I ($C_1 \perp \! \! \! \perp C_2$)}: $\Sigma_k = 10^{-2} I$ and $m_k \sim \mathcal{U}([-1, 1]^5)$ for all $k$, and $\sigma = 0.1$.  
$A = (a_{ij})$ is diagonal with $a_{ii} = 0.1/i$.

\noindent\textbf{Setting II ($C_1 \not \perp \! \! \! \perp C_2$)}: $\Sigma_k = I$ and $m_k \sim \mathcal{U}([-5, 5]^5)$ for all $k$, and $\sigma = 1$.  
$A = (a_{ij})$ with $a_{ii} = 0.5$ and $a_{ij} = 0.25$ for $i \ne j$.

\begin{table*}[th!]
\centering
\caption{Means and standard deviations of 2-Wasserstein ($W_2$) distances of each methods (over 50 repeats) under Setting I and II with 1000 samples. All relevant parameters are tuned to minimize the $W_2$ distance. 
The best results are highlighted in bold, the second-best results are underlined, and the third-best results are dash-underlined.}
\label{tab_setting_1}
\begin{tabular}{lcccc}
\toprule
\multirow{2}{*}{\textbf{Method}} & \multicolumn{2}{c}{\textbf{Setting I}} & \multicolumn{2}{c}{\textbf{Setting II}} \\ \cmidrule{2-5}
& $W_2$ ($\times 10^2$) ($\downarrow$) & Guidance scales & $W_2$ ($\downarrow$) & Guidance scales \\
\midrule
DMDG (\ref{eq_double_guidance_approx})  & 3.032(1.568) & $\lambda_1=\lambda_2=180$ & \dashuline{3.275}(3.411) & $\lambda_1=\lambda_2=1.5$ \\
DMIDG & 3.095(1.646) & $\lambda_1=\lambda_2=180$ & 4.688(6.447) & $\lambda_1=\lambda_2=2$ \\
DMHG (\ref{eq_hybrid_guidance}) & \textbf{2.097}(0.830) & $\lambda_1=1, \lambda_2=250$ & \textbf{2.386}(2.923) & $\lambda_1=\lambda_2=1.5$ \\
DMIHG & \underline{2.162}(1.106) & $\lambda_1=1, \lambda_2=250$ & 4.919(6.415) & $\lambda_1=\lambda_2=1.5$ \\
Composition & 23.452(16.471)  &  $\lambda_1=\lambda_2=1$ & 22.432(10.369) &  $\lambda_1=\lambda_2=1$ \\
COIND$^*$ &  \dashuline{2.839}(1.049) & N/A &  \underline{2.746}(2.993) & N/A \\
CTRL & 4.057(2.949) & N/A & 5.620(6.045) & N/A \\
Regressor & 157.645(94.224) & N/A & 6.800(3.818) & N/A \\
GAIN & 26.312(14.977) & N/A & 6.169(2.138) & N/A \\
Forest Diffusion & 17.639(18.648) & N/A & 8.210(6.747) & N/A \\
KNN & 5.591(4.790) & N/A & 5.469(4.655) & N/A \\
\bottomrule
\end{tabular}

N/A: Guidance scales are not available; $^*$: COIND is trained on \textcolor{red}{full} datasets without missing values. 
\end{table*}

The results are summarized in Table \ref{tab_setting_1}. In both Settings I and II, DMHG and DMIHG exhibit clear advantages, indicating that the CFG-based methods are more effective at matching the target conditional distribution (except COIND, which is trained on full datasets). The CG-based methods, DMDG and DMIDG, perform slightly worse than their CFG-based counterparts, which is consistent with findings reported in prior work \cite{ho_classifier_free,Wu_2024_big_scale_distort_cond_dist}. In addition, CG-based methods consistently outperform imputation-based and RL-based approaches.
In Setting I, due to the independence of $C_1$ and $C_2$, 
DMDG and DMHG perform similarly to their counterparts DMIDG and DMIHG, although our methods exhibit lower variances.
In Setting II, the structure of matrix $A$ induces dependency between $C_1$ and $C_2$, leading to a substantial performance gap: both DMDG and DMHG significantly outperform DMIDG and DMIHG. This observation is supported by the phenomenon shown in right panel of Figure \ref{fig_vector_fields} that when $C_1 \not \perp \! \! \! \perp C_2$, DMIDG introduces errors in the score estimation. The performances of Composition are not ideal due to the violation of conditional independence, as we mentioned in Appendix \ref{sec_compared_methods}. 
In terms of parameter selection, our parameters are consistent in order of magnitude with the theoretical analysis presented in Theorem \ref{thm_square_of_densities}, i.e. $\lambda_1,\lambda_2 \sim \mathcal{O}(1/\sigma^2)$.

\section{More General Missing Type and Diffusion Model with Triple Guidance (DMTG)}
\label{sec_dmtg}

\begin{table*}[htbp!]
\centering
\caption{Illustration of more general block-wise missing patterns.}
\label{table_block_wise_missing_combined}

\caption*{\textbf{(a) Missing Type I} (overlapping conditions)}
\begin{tabular}{|c|c|c|c|c|}
\hline
Dataset & Target & Condition 1 & Condition 2 & Condition 3 \\ \hline
$D^{(1)}$ & $X_0^{(1)}$ & $C_1^{(1)}$ & $\varnothing$ & $C_3^{(1)}$ \\ \hline
$D^{(2)}$ & $X_0^{(2)}$ & $\varnothing$ & $C_2^{(2)}$ & $C_3^{(2)}$ \\ \hline
\end{tabular}

\vspace{1.0em}

\caption*{\textbf{(b) Missing Type II} (three datasets with non-overlapping conditions)}
\begin{tabular}{|c|c|c|c|c|}
\hline
Dataset & Target & Condition 1 & Condition 2 & Condition 3 \\ \hline
$D^{(1)}$ & $X_0^{(1)}$ & $C_1^{(1)}$ & $\varnothing$ & $\varnothing$ \\ \hline
$D^{(2)}$ & $X_0^{(2)}$ & $\varnothing$ & $C_2^{(2)}$ & $\varnothing$ \\ \hline
$D^{(3)}$ & $X_0^{(3)}$ & $\varnothing$ & $\varnothing$ & $C_3^{(3)}$ \\ \hline
\end{tabular}
\end{table*}

Our double guidance methods can be extended to process more aggregated datasets and missing conditions. Table \ref{table_block_wise_missing_combined} presents the two additional missing types: \textbf{missing type I}, where overlapping conditions exist, and \textbf{missing type II}, which has three aggregated datasets with non-overlapping conditions.

For missing type I, we can approximate the true conditional score $\nabla \log p_t(X_t|C_1, C_2, C_3)$ by:
\begin{align}
\label{eq_triple_guidance_intersec_cg}
    & \nabla \log p_t(X_t|C_1, C_2, C_3) \notag \\ 
    & =  \nabla \log p_t(X_t|C_3) + \nabla \log p(C_1|X_t,C_3) + \nabla \log p(C_2|X_t,C_1,C_3) \notag \\
    &\approx s_{\theta}(X_t,t,\varnothing, C_3) - \lambda_1 \nabla \| C_1 - \widehat{f}_1(X_{0|t,C_3})  \|_2^2  - \lambda_{2} \nabla \| C_2 - \widehat{f}_{2} (X_{0|t,C_1,C_3})   \|_2^2,  
\end{align}
or
\begin{align}
\label{eq_triple_guidance_intersec_cfg}
    & \nabla \log p_t(X_t|C_1, C_2, C_3) \notag \\
    &\approx (1-\lambda_{1}) s_{\theta}(X_t,t,\varnothing,C_3) + \lambda_{1} s_{\theta}(X_t,t,C_1,C_3)  - \lambda_2 \nabla \| C_2 - \widehat{f}_2(X_{0|t,C_1,C_3})  \|_2^2,
\end{align}
where the neural network $nn_{\theta}$ now have four inputs: $X_t,t,C_1$ and $C_3$. The latter two can be masked if unavailable. Accordingly, we define $X_{0|t,C_3} := nn_{\theta}(X_t,t,\varnothing,C_3)$, $X_{0|t,C_1,C_3} := nn_{\theta}(X_t,t,C_1,C_3)$, $s_{\theta}(X_t,t,\varnothing,C_3):=[nn_{\theta}(X_t,t,\varnothing,C_3)-X_t]/t^2$ and $ s_{\theta}(X_t,t,C_1,C_3):=[nn_{\theta}(X_t,t,C_1,C_3)-X_t]/t^2$. The derivations of (\ref{eq_triple_guidance_intersec_cg}) and (\ref{eq_triple_guidance_intersec_cfg}) are similar to (\ref{eq_our_cond_score_appendix}). Since the joint samples of $(X_0, C_1,C_3)$ exist, $X_{0|t,C_3}$ and $X_{0|t,C_1,C_3}$ are learnable. Conceptually, they are equivalent to DMDG and DMHG, with the only difference being the inclusion of $C_3$ in the network input. Therefore, we refer to (\ref{eq_triple_guidance_intersec_cg}) and (\ref{eq_triple_guidance_intersec_cfg}) as \textbf{DMDG-I} and \textbf{DMHG-I}, respectively.
By replacing $X_{0|t, C_1, C_3}$ with $X_{0|t, C_3}$, we obtain DMIDG-I and DMIHG-I.

We can also apply our method to more extreme block-wise missing datasets. For the dataset with missing type II illustrated in Table \ref{table_block_wise_missing_combined} \textbf{(b)}, the conditional score function can be decomposed as:
\begin{align}
\label{eq_triple_guidance_3dataset}
    &\nabla \log p_t(X_t|C_1, C_2, C_3) \notag \\
    &= \nabla \log p(X_t|C_1, C_2) + \nabla \log \int p(C_3|X_0)p(X_0|X_t,C_1,C_2) d X_0 \notag \\
    &\approx \nabla \log p(X_t|C_1, C_2) + \nabla \log p(C_3|\mathbb{E}\left[ X_0 |X_t,C_1,C_2 \right])
\end{align}
In (\ref{eq_triple_guidance_3dataset}), DMDG (\ref{eq_double_guidance_approx}) or DMHG (\ref{eq_hybrid_guidance}) can be directly used to approximate $\nabla \log p(X_t|C_1, C_2)$. The challenge of this missing type comes from how to estimate the posterior mean $ \mathbb{E}\left[ X_0 |X_t,C_1,C_2 \right]$. Our solution originates from the Tweedie projection~(\ref{eq_tweedie_proj}):
\begin{equation}
    \mathbb{E}[X_0 | X_t, C_1, C_2]
    = X_t + t^2 \cdot \nabla \log p_t(X_t | C_1, C_2),
\end{equation}
where $\nabla \log p_t(X_t | C_1, C_2)$ can be directly estimated using
DMDG~(\ref{eq_double_guidance_approx}) or DMHG~(\ref{eq_hybrid_guidance}).
This yields the following estimations of $\mathbb{E}[X_0 | X_t, C_1, C_2]$:
\begin{align}
\label{eq_dmtg_tweedie_proj}
    X^{\text{DMDG}}_{0|t,C_1,C_2}
    &:= X_{0|t}
    - \lambda_1 t^2 \nabla \| C_1 - \widehat{f}_1(X_{0|t}) \|_2^2
    - \lambda_2 t^2 \nabla \| C_2 - \widehat{f}_2(X_{0|t,C_1}) \|_2^2, \notag \\
    X^{\text{DMHG}}_{0|t,C_1,C_2}
    &:= (1-\lambda_1) X_{0|t}
    + \lambda_1 X_{0|t,C_1}
    - \lambda_2 t^2 \nabla \| C_2 - \widehat{f}_2(X_{0|t,C_1}) \|_2^2.
\end{align}
These two estimates can also be interpreted as performing two gradient descent from $X_{0 | t}$, and this estimation strategy can potentially be extended to more aggregated datasets. Then, (\ref{eq_triple_guidance_3dataset}) can be estimated by:
\begin{align}
\label{eq_dmtg_2}
    &\nabla \log p_t(X_t|C_1, C_2, C_3) \notag \\
    & \approx s_{\theta}(X_t, t, \varnothing) - \lambda_1 \nabla \|C_1-\widehat{f}_1(X_{0|t}) \|_2^2  - \lambda_2 \nabla \|C_2-\widehat{f}_2(X_{0|t,C_1}) \|_2^2 - \lambda_3 \nabla \|C_3-\widehat{f}_3(X^{\text{DMDG}}_{0|t,C_1,C_2}) \|_2^2, 
\end{align}
or
\begin{align}
\label{eq_dmthg_2}
    & \nabla \log p_t(X_t|C_1, C_2, C_3) \notag \\
    & \approx (1-\lambda_1)s_{\theta}(X_t, t, \varnothing) + \lambda_1 s_{\theta}(X_t, t, C_1) - \lambda_2 \nabla \|C_2-\widehat{f}_2(X_{0|t,C_1}) \|_2^2 - \lambda_3 \nabla \|C_3-\widehat{f}_3(X^{\text{DMHG}}_{0|t,C_1,C_2}) \|_2^2.
\end{align}
We refer to (\ref{eq_dmtg_2}) as \textbf{Diffusion Model with Triple Guidance (DMTG)}, and (\ref{eq_dmthg_2}) as \textbf{Diffusion Model with Triple Hybrid Guidance (DMTHG)}. Replacing $X_{0|t,C_1}$, $X^{\text{DMDG}}_{0|t,C_1,C_2}$ and $X^{\text{DMHG}}_{0|t,C_1,C_2}$ with $X_{0|t}$ results in Diffusion Model with Independent Triple Guidance (DMITG) and Diffusion Model with Independent Triple Hybrid Guidance (DMITHG). However, the estimations of $\mathbb{E}[X_0|X_t,C_1,C_2]$ (\ref{eq_dmtg_tweedie_proj}) introduce a serious issue: when the scale of $t$ is large, these estimations can lead to numerical explosion.
To address this, we apply the guidance of $C_3$ only when $t \leq 1$, which effectively prevents numerical instability while still enabling control over $C_3$. Consequently, in our simulations of DMTG and DMTHG, the guidance scale $\lambda_3$ is larger than $\lambda_1$ and $\lambda_2$.

We conduct simulations based on the data generation mechanism from Setting II in Appendix \ref{sec_app_gaussian_mix} to evaluate the performance of DMDG-I, DMHG-I, DMTG and DMTHG. For missing type I, the sample sizes of $D^{(1)}$ and $D^{(2)}$ are all set to 10,000. For missing type II, sample sizes of $D^{(1)}$, $D^{(2)}$ and $D^{(3)}$ are 6,667. We define the condition variables as follows: $C_1 = (Y_1, Y_2)$, $C_2 = (Y_3, Y_4)$, and $C_3 = Y_5$. We compare guidance-based methods and imputation-based methods. CTRL can only handle two conditions so it is no longer suitable for triple conditions.

The results can be found in Table \ref{tab_dmtg_split_vertical} and it should be interpreted in conjunction with Table \ref{tab_setting_1}. We observe that the datasets under missing type I (Table \ref{table_block_wise_missing_combined}, (a)) exhibit the lowest level of missingness, while the datasets in Table \ref{table_block_wise_missing} present a moderate degree, and the missing type II (Table \ref{table_block_wise_missing_combined}, (b)) shows the highest level of missingness. Correspondingly, our proposed methods (DMDG, DMHG, DMDG-I, DMHG-I, DMTG, DMTHG), as well as imputation-based baselines, achieve lower $W_2$ distances under missing type I, moderate distances on the datasets in Table \ref{table_block_wise_missing}, and the highest $W_2$ distances under missing type II. However, methods of independent guidance perform relatively poorly under missing type II, as they do not take into account CI. Notably, DMTG and DMTHG remain effective, though the guidance of $C_3$ only exists when $t \leq 1$. 

\begin{table*}[htbp!]
\centering
\caption{$W_2$ distances and standard deviations of each methods under different missing types. All experiments are repeated 50 times and evaluated on 1000 samples. All parameters are tuned to minimize the $W_2$ distance. The best results are highlighted in bold, and the second-best results are underlined.}
\label{tab_dmtg_split_vertical}

\caption*{\textbf{(a) Missing type I} (overlapping conditions)}
\begin{tabular}{lcc}
\toprule
\textbf{Method} & $W_2$ ($\downarrow$) & Guidance scales \\
\midrule
DMDG-I (\ref{eq_triple_guidance_intersec_cg})  &  \underline{3.056}(2.740) & $\lambda_1=\lambda_2=1.5$ \\
DMIDG-I  & 5.251(5.625) & $\lambda_1=\lambda_2=1.5$ \\
DMHG-I (\ref{eq_triple_guidance_intersec_cfg})  & \textbf{2.147}(1.537) & $\lambda_1=\lambda_2=1.5$ \\
DMIHG-I & 4.279(4.023) & $\lambda_1=\lambda_2=1.5$ \\
Regressor & 5.136(4.696) & N/A \\
GAIN & 5.467(3.879) & N/A \\
Forest Diffusion & 5.854(5.067) & N/A \\
KNN & 4.853(4.375) & N/A \\
\bottomrule
\end{tabular}

\vspace{1.0em}

\caption*{\textbf{(b) Missing type II} (three datasets)}
\begin{tabular}{lcc}
\toprule
\textbf{Method} & $W_2$ ($\downarrow$) & Guidance scales \\
\midrule
DMTG (\ref{eq_double_guidance_approx})  & \underline{3.857}(2.815) & $\lambda_1=\lambda_2=1.5,\lambda_3=7$ \\
DMITG  & 14.460(20.595) & $\lambda_1=\lambda_2=1.5,\lambda_3=7$ \\
DMTHG (\ref{eq_hybrid_guidance}) & \textbf{3.808}(3.546) & $\lambda_1=\lambda_2=1.5,\lambda_3=7$\\
DMITHG & 4.819(5.887) & $\lambda_1=\lambda_2=1.5,\lambda_3=7$ \\
Regressor &9.122(4.750) & N/A \\
GAIN & 16.623(14.677) & N/A \\
Forest Diffusion & 12.760(10.391) & N/A \\
KNN & 6.472(3.794) & N/A \\
\bottomrule
\end{tabular}
\end{table*}

\section{Applying DMDG and DMHG to Other Diffusion Models}
To demonstrate the generality of DMDG and DMHG, we apply them to several classic diffusion models, including Denoising Diffusion Probabilistic Models \citep[DDPM,][]{ho_ddpm} and Score-Based Generative Models \citep[SGM,][]{song_sgm}. 

The data generation procedure follows Setting II in Appendix \ref{sec_app_gaussian_mix}. For the DDPM model, we adopt the default settings in \cite{ho_ddpm}. The SGM model follows the default configuration in \cite{yang_2024_cdcit}. The EDM model uses the same settings as in the main text, except that we set the sampling steps to 1000 in order to match the number of sampling steps used in DDPM and SGM.

The results can be found in Table~\ref{tab_different_diffusions}. As shown, DMDG and DMHG are also applicable to both DDPM and SGM. A consistent observation across all models is that DMDG outperforms DMIDG, and DMHG outperforms DMIHG.

\begin{table*}[ht]
\centering
\caption{Means and standard deviations of $W_2$ distances of each diffusion models and methods (over 50 repeats) under Setting II with 1000 samples. All relevant parameters are tuned to minimize the $W_2$ distance. 
The best results are highlighted in bold, and the second-best results are underlined.}
\label{tab_different_diffusions} 

\begin{tabular}{llcc}
\toprule
\textbf{Diffusion model} & \textbf{Method} & $W_2$ ($\downarrow$) & Guidance scales \\
\midrule
EDM \cite{karras_edm} &DMDG (\ref{eq_double_guidance_approx})  & \underline{3.059}(3.573)  & $\lambda_1=\lambda_2=1.5$ \\
&DMIDG  & 4.604(6.523) & $\lambda_1=\lambda_2=2$ \\
&DMHG (\ref{eq_hybrid_guidance}) & \textbf{2.004}(2.257) & $\lambda_1=\lambda_2=1.5$\\
&DMIHG & 4.128(5.503) & $\lambda_1=\lambda_2=1.5$ \\
\hline

DDPM \cite{ho_ddpm} &DMDG (\ref{eq_double_guidance_approx})  & \underline{3.498}(4.835)  & $\lambda_1=\lambda_2=6$ \\
&DMIDG  & 6.189(10.586) & $\lambda_1=\lambda_2=6$ \\
&DMHG (\ref{eq_hybrid_guidance}) & \textbf{1.894}(1.805) & $\lambda_1=1, \lambda_2=6$\\
&DMIHG & 5.026(7.640) & $\lambda_1=1, \lambda_2=6$ \\
\hline

SGM \cite{song_sgm} &DMDG (\ref{eq_double_guidance_approx})  & 3.878(3.156)  & $\lambda_1=\lambda_2=4.5$ \\
&DMIDG  & 4.915(4.379) & $\lambda_1=\lambda_2=4.5$ \\
&DMHG (\ref{eq_hybrid_guidance}) & \textbf{2.037}(0.920) & $\lambda_1=1, \lambda_2=4.5$\\
&DMIHG & \underline{3.248}(2.534) & $\lambda_1=1, \lambda_2=4.5$ \\
\bottomrule
\end{tabular}

\end{table*}

\section{Compared Methods in Detail}
\label{sec_compared_methods}
\subsection{DMIDG and DMIHG}
\label{sec_app_dmihg}
Unlike DMDG and DMHG, DMIDG and DMIHG model the joint density $p(C_1, C_2 | X_t)$ as the product of two independent densities, $p(C_1 | X_t)$ and $p(C_2 | X_t)$.
This simplification ignores the fact that $C_1$ and $C_2$ are generally not conditionally independent given $X_t$, leading to a biased estimation of the true conditional score, as shown in Figure \ref{fig_vector_fields}. The deteriorated performances of DMIDG and DMIHG can be observed in Figure \ref{fig_molecular_task_12} and Table \ref{tab_setting_1}.

Algorithmically speaking, these methods replace the dependency-aware density $p(C_2|X_t,C_1)$ in (\ref{eq_our_cond_score_appendix})  with $p(C_2|X_t)$, i.e.:
\begin{align}
    &\nabla \log p_t(X_t|C_1, C_2) \notag \\
    & \not = \nabla \log p_t(X_t) +\nabla \log p(C_1|X_t)  + \nabla \log p(C_2|X_t) \notag \\
    & \approx \nabla \log p_t(X_t) + \nabla \log p(C_1|\mathbb{E} \left[ X_0|X_t\right] )  + \nabla \log p(C_2|\mathbb{E}\left[X_0|X_t\right])  \notag \\
    & \approx s_{\theta}(X_t, t, \varnothing)  - \lambda_1 \nabla \|C_1-\widehat{f}_1(X_{0|t}) \|_2^2  - \lambda_2 \nabla \|C_2-\widehat{f}_2(X_{0|t}) \|_2^2 \label{eq_dmidg} \\
    & \approx (1-\lambda_1)s_{\theta}(X_t, t, \varnothing) + \lambda_1 s_{\theta}(X_t, t, C_1)   - \lambda_2 \nabla \|C_2-\widehat{f}_2(X_{0|t}) \|_2^2, \label{eq_dmihg}
\end{align}
where (\ref{eq_dmidg}) and (\ref{eq_dmihg}) denote the approximate conditional scores of DMIDG and DMIHG, respectively. 

To better illustrate the importance of conditional independence, we use conditional mutual information (CMI) to visualize how the relationship between $C_1$ and $C_2$ evolves over $t$, conditioned on different variables. The CMI, denoted as $I(C_1; C_2 | X_t)$, can be interpreted as a measure of the conditional dependence between $C_1$ and $C_2$ given $X_t$. $I(C_1; C_2 | X_t) = 0$ implies $C_1 \perp \!\!\!\perp C_2 | X_t$. Conversely, $I(C_1; C_2 | X_t) > 0$ means $C_1 \not \perp \!\!\!\perp C_2 | X_t$ 
\cite{shuai_2022_nnscit,shuai_2023_nnlscit}. 
It can be observed in Figure \ref{fig_CI_relationship} that as $t$ increases, both $I(C_1; C_2 | X_t)$ and $I(C_1; C_2 | X_{0|t})$ increase accordingly, indicating $C_1 \not \perp \! \! \! \perp C_2 |X_{t}$ and $C_1 \not \perp \! \! \! \perp C_2 | X_{0|t}$ when $t$ becomes large. In contrast, $I(C_1; C_2 | X_{0|t,C_1})$ remains close to zero across all values of $t$, since $C_1$ is conditioned.

\subsection{Compositional methods}
Compositional methods aim to combine different score functions to achieve conditional composition. Under the assumption of $C_1 \perp \! \! \! \perp C_1$, \citet{Liu_2022_compositional_diff} proposed to decompose the score function by:
\begin{align}
    & \nabla \log p_t(X_t|C_1, C_2) \notag \\
    &  = \nabla \log p_t(X_t) + \nabla \log p(C_1,C_2|X_t) \notag \\
    & \not = \nabla \log p_t(X_t) + \nabla \log p(C_1|X_t) + \nabla \log p(C_2|X_t) \label{eq_why_composition_fail} \\
    & =  \nabla \log p_t(X_t) + \nabla \log \left[ p(X_t|C_1)/p(X_t) \right]  + \nabla \log \left[ p(X_t|C_2)/p(X_t) \right] \notag \\ 
    & \approx s_{\theta}(X_t,t,\varnothing, \varnothing)  + \lambda_1 \left[ s_{\theta}(X_t,t,C_1, \varnothing) - s_{\theta}(X_t,t,\varnothing, \varnothing) \right]  + \lambda_2 \left[ s_{\theta}(X_t,t, \varnothing,C_2) - s_{\theta}(X_t,t,\varnothing, \varnothing) \right].
\end{align}
In (\ref{eq_why_composition_fail}), it is assumed that $C_1 \perp \! \! \! \perp C_2 |X_t$. However, as shown in Figure~\ref{fig_CI_relationship}, $C_1$ and $C_2$ are not conditionally independent given $X_t$. The bias in the estimation of score function was also observed by \citet{gaudi_2025_coind}.

\subsection{COIND}
COIND is a principled and elegant method capable of logically decomposing and combining multiple conditions~\cite{gaudi_2025_coind}.
This is achieved by adding an additional penalty term $L_{\text{penalty}}$ to the training loss:
\begin{align}
    &L_{\text{penalty}}  \notag \\
    & \quad = \mathbb{E}_{X_t,C_1,C_2} \| s_{\theta}(X_t,t,\varnothing, \varnothing)  +  \left[ s_{\theta}(X_t,t,C_1, \varnothing) - s_{\theta}(X_t,t,\varnothing, \varnothing) \right]  + \left[ s_{\theta}(X_t,t, \varnothing,C_2) - s_{\theta}(X_t,t,\varnothing, \varnothing) \right]  \notag \\
    &  \quad \quad \quad \quad  \quad \quad  \quad - s_{\theta}(X_t,t,C_1, C_2) \|_2^2,
\end{align}
which effectively enforces $s_{\theta}(X_t,t,C_1, C_2)$
to be consistent with the assumption that
$C_1 \perp\!\!\!\perp C_2 | X_t$, as shown in (\ref{eq_why_composition_fail}).

Although practical, such a penalty could introduce bias in $s_{\theta}$, as shown in the left panel of Figure \ref{fig_CI_relationship} and Figure \ref{fig_vector_fields}. We argue that, as in DMDG and DMHG,
directly injecting a $C_1$-aware guidance of $C_2$
provides a unbiased score estimation without violating the assumption
$C_1 \not\!\perp\!\!\!\perp C_2 | X_t$. In addition, COIND requires access to the complete dataset during training due to the appearance of $s_{\theta}(X_t,t,C_1, C_2)$.
As a result, we are only able to include comparisons with COIND
in the simulation study presented in Appendix~\ref{sec_app_gaussian_mix},
and cannot apply it to real-world datasets.

\subsection{Conditioning pre-Trained diffusion models with Reinforcement Learning (CTRL)}
\citet{zhao_2025_adding_cond_by_rl} proposed incorporating additional conditions into a pre-trained CDM via reinforcement learning. Their method involves fine-tuning a pre-trained neural network $nn_{\theta} \left(X_t, t, C_1 \right)$. Let $nn_{\theta,\psi}(X_t,t,C_1,C_2)$ denote the neural network subject to fine-tuning, where $\theta$ represents the original parameters and $\psi$ denotes the parameters for processing $C_2$, extended by Controlnet \cite{zhang_2023_control_net}. The loss function used to fine-tune the CDM consists of two components:
\begin{align}
\label{eq_rl_funetune_loss}
    &\int_{t_{\min}}^{t_{\max}} \dfrac{\| nn_{\theta} \left(X_t, t, C_1 \right) - nn_{\theta,\psi}(X_t,t,C_1,C_2) \|_2^2}{2t^3} dt -\gamma \log p(C_2|X_{t_{\min}},C_1)
\end{align} 
where $\gamma$ is a fixed guidance scale. The first component is the weighted discrepancy between the trajectories of the original diffusion model and the fine-tuned diffusion model. The second component is the negative log-likelihood, or in terms of RL, the reward function. $X_{t_{\min}}$ is the generated sample at the terminate of the reverse process.  By minimizing (\ref{eq_rl_funetune_loss}), the fine-tuned model is capable of generating samples conditioned on $C_1$ and $C_2$. When $C_1$ and $C_2$ are conditionally independent given $X_0$, the negative log-likelihood term can be simplified to $-\gamma \log p(C_2|X_{t_{\min}})$.

Although this fine-tuning approach is indeed well-motivated, it still has several drawbacks. First, as a fine-tuning technique, it depends on a sufficiently strong pre-trained model. In our setting, this implies that one should first train $nn_{\theta}(X_t,t,C_1)$ on a $D^{(1)}$ with a huge sample size, then fine-tune $nn_{\theta, \psi}$ on $D^{(2)}$. In practice, such $D^{(1)}$ seldom exists. Second, the fine-tuning method was originally designed for datasets such as $D^{(2)} = \{ X_0^{(2)}, C_1^{(2)}, C_2^{(2)}  \}$ and the pair $(C_1,C_2)$ explicitly appears in (\ref{eq_rl_funetune_loss}). In the case of (\ref{eq_function_map}), \cite{zhao_2025_adding_cond_by_rl} proposed using $C_1 \perp \!\!\! \perp C_2 |X_0$ to eliminate $C_1$ in the negative log-likelihood, and uniformly sampling $C_1$ then coupling it with samples in $C_2^{(2)}$. In other words, $(C_1,C_2)$ is drawn from $P_{C_1} \times P_{C_2}$ rather than $P_{C_1,C_2}$. Such replacement may undermine the fine-tuning of $nn_{\theta,\psi}(X_t,t,C_1,C_2)$ when miss-matched $(C_1,C_2)$ appears. Furthermore, the reward function is the negative log-likelihood only at the end of the reverse process (sparsity), instead of the entire reverse process, which may impair the conditional control. Finally, the scale of negative log-likelihood $\gamma$, is predetermined during training, preventing flexible adjustment at generation time.

\subsection{Imputation methods}
We compared regressor imputation, GAIN \cite{Yoon_2018_gain}, Forest diffusion \cite{Jolicoeur_2024_forestdiffusion} and KNN imputation. Regressor imputation is the simplest approach, where the missing $C_2^{(1)}$ and $C_1^{(2)}$ are replaced by $\widehat{f}_2(X_0^{(1)})$ and $\widehat{f}_1(X_0^{(2)})$. However, this replacement neglects the stochastic nature of $C_2^{(1)}$ and $C_1^{(2)}$ because $\widehat{f}_1$ and $\widehat{f}_2$ only estimate the conditional mean (see (\ref{eq_function_map})). GAIN and Forest Diffusion are both generative-model-based imputation algorithms. The former leverages generative adversarial networks (GANs), while the latter is built upon diffusion models. Although both methods are capable of capturing the stochasticity of $C_1$ and $C_2$, their performance is limited due to the structure of block-wise missingness. In the experiments presented in Appendix \ref{sec_app_gaussian_mix}, the best imputation method is KNN. Due to the block-wise missing structure, KNN depends solely on information from $X_0$, which aligns with the data generation mechanism described in (\ref{eq_function_map}). However, most imputation-based approaches still perform worse than guidance methods and CTRL.

\section{Additional Experimental Details and Results}
\label{sec_additional_exps}

\subsection{Calculating $W_2$ distance and extra results of De novo drug design}
\label{sec_app_design_mole_from_0}

In section \ref{sec_design_mole_from_0}, we adopt COATI, a powerful variational autoencoder (VAE), to encode molecules into a latent space and decode latent vectors back into molecules \cite{Kaufman_2024_COATI}. The diffusion models then operate within the latent space. 
To evaluate how well the generated molecules align with the target conditional distributions, we construct two reference sets.
We first select 500 molecules from GEOM-DRUG whose properties $C_1$ are not only closest to the target $C_1$  (in terms of $\ell_2$ distance), but also satisfy the requirements of Task 1.
Similarly, we select 500 molecules from ZINC250k whose properties $C_2$ are closest to the target $C_2$ and satisfy the criteria of Task 2.
We then compute the $W_2$ distance between the latent vectors of the generated molecules and those of the corresponding reference molecules. The results are shown in Figure \ref{fig_molecular_task_12}.

We provide the detailed success rates for Task 1 and Task 2 (\ref{eq_molecular_task_12}) shown in Figure \ref{fig_molecular_task_12}, paired with the corresponding guidance scales. The results are summarized in Table \ref{tab_molecular_task_12}. It can be observed that, compared to DMIDG, DMDG achieves a higher task success rate with smaller guidance scales, which also is the key to its lower $W_2$ distance to the GEOM dataset.

\begin{table*}[ht]
  \centering
  \caption{Detailed success rates and corresponding guidance scales used for each method in Figure \ref{fig_molecular_task_12}.}
  \label{tab_molecular_task_12}

  \begin{subtable}[t]{0.48\textwidth}
    \centering
    \caption{DMDG}
    \begin{tabular}{ccccc}
      \toprule
      Task 1\&2 & Task 1 & Task 2 & $\lambda_1$ & $\lambda_2$ \\
      \midrule
       0.1  \% & 0.9 \%  & 49.9 \% & 0 & 0 \\
       12.6 \% & 26.1 \%  & 52.6 \% & 6 & 3 \\
       21.3 \% & 34.6 \%  & 68.3 \% & 9 & 6 \\
       30.5 \% & 51.5 \%  & 63.8 \% & 12 & 6 \\
       39.7 \% & 78.0 \%  & 54.5 \% & 21 & 6 \\
       50.5 \% & 81.2 \%  & 64.6 \% & 27 & 9 \\
       60.7 \% & 89.5 \%  & 68.3 \% & 42 & 12 \\
       70.4 \% & 93.2 \%  & 75.5 \% & 60 & 18 \\
       76.0 \% & 93.4 \%  & 81.5 \% & 63 & 21 \\
      \bottomrule
    \end{tabular}
  \end{subtable}%
  \hfill%
  \begin{subtable}[t]{0.48\textwidth}
    \centering
    \caption{DMIDG}
    \begin{tabular}{ccccc}
      \toprule
      Task 1\&2 & Task 1 & Task 2 & $\lambda_1$ & $\lambda_2$\\
      \midrule
       0.1  \%  & 0.9 \% & 49.9 \% & 0 & 0  \\
       10.5 \% & 12.5 \% & 98.5 \% & 9 & 27 \\
       21.1 \% & 21.5 \% & 97.7 \% & 18 & 33 \\
       30.4 \% & 31.5 \% & 96.6 \% & 24 & 27 \\
       40.6 \% & 43.2 \% & 95.0 \% & 33 & 30 \\
       50.1 \% & 55.2 \% & 92.4 \% & 51 & 33 \\
       55.2 \% & 61.7 \% & 91.9 \% & 57 & 30 \\
       58.7 \% & 66.2 \% & 90.6 \% & 63 & 30 \\
       N/A & N/A & N/A & N/A & N/A \\
      \bottomrule
    \end{tabular}
  \end{subtable}

  \vspace{1em}

  \begin{subtable}[t]{0.48\textwidth}
    \centering
    \caption{DMHG}
    \begin{tabular}{ccccc}
      \toprule
      Task 1\&2 & Task 1 & Task 2 &  $\lambda_1$ & $\lambda_2$ \\
      \midrule
      0.1 \% & 0.9 \% & 49.9 \% & 0 & 0 \\
      10.3 \% & 22.9 \% & 82.1 \% & 3 & 3 \\
      15.9 \% & 28.3 \% & 67.2 \% & 7 & 6 \\
      N/A & N/A & N/A & N/A & N/A \\
      \bottomrule
    \end{tabular}
  \end{subtable}%
  \hfill%
  \begin{subtable}[t]{0.48\textwidth}
    \centering
    \caption{DMIHG}
    \begin{tabular}{ccccc}
      \toprule
      Task 1\&2 & Task 1 & Task 2 & $\lambda_1$ & $\lambda_2$ \\
      \midrule
       0.1 \% & 0.9 \% & 49.9 \% & 0 & 0 \\
       9.9 \% &  26.9 \% & 43.4 \% & 3 & 3 \\
       15.1 \% & 28.2 \% & 65.1 \% & 5 & 9 \\
       17.3 \% & 26.3 \% & 68.5 \% & 8 & 15 \\
      \bottomrule
    \end{tabular}
  \end{subtable}

\end{table*}

\subsection{Additional metrics for generated molecules of De novo drug design}
In Table \ref{tab_valid_unique_similar}, we report several widely used metrics for evaluating generated molecules, including Validity, Novelty, and Diversity \cite{Haote_Li_KAE,SY_chemguide,hoogeboom_2022_e3_diffusion,Gebauer_2022_cond_drug_design_1}. Validity refers to the proportion of generated molecules that are chemically valid. Novelty measures the proportion of generated molecules that are not in the training set. Diversity is defined as follows:
\begin{equation*}
    \text{Diversity} = 1 - \dfrac{2}{n(n-1)} \sum_{\text{Mol}_1, \text{Mol}_2} \text{Sim}(\text{Mol}_1,\text{Mol}_2),
\end{equation*}
where Sim denotes the Tanimoto similarity, and $\text{Mol}_1$ and $\text{Mol}_2$ refer to two generated molecules in one batch.

\begin{table*}[htbp]
\centering
\caption{Every metric is evaluated on a batch with 1000 generated molecules, repeated by 5 times.}
\label{tab_valid_unique_similar}
\begin{tabular}{ccccc}
\toprule
\textbf{Method} & \textbf{Guidance scales} & \textbf{Validity} (\%, $\uparrow$) & \textbf{Uniqueness} (\%, $\uparrow$)& \textbf{Diversity} ($\uparrow$)\\
\midrule
DMDG & $\lambda_1=63,\lambda_2=21$ & 99.440(0.351) & 99.840(0.114) & 0.851(0.001) \\

DMIDG & $\lambda_1=63,\lambda_2=30$ & 99.200(0.228) & 98.660(0.571) & 0.853(0.001) \\

DMHG & $\lambda_1=7, \lambda_2=6$ & 99.620(0.203) & 95.780(1.907) & 0.836(0.001) \\

DMIHG & $\lambda_1=8,\lambda_2=15$ & 99.680(0.193) & 96.000(1.596) & 0.844(0.001) \\
\hline
Uncond & $\lambda_1=\lambda_2=0$ & 99.660(0.167) & 99.620(0.228) & 0.881(0.001) \\

\bottomrule
\end{tabular}
\end{table*}

\subsection{Predicting Tanimoto similarity and details of partial diffusion}
\label{sec_app_tanimoto_regressor}
\begin{figure}[ht!]
    \centering
    \begin{subfigure}[t]{0.45\textwidth}
        \centering
        \includegraphics[width=\linewidth]{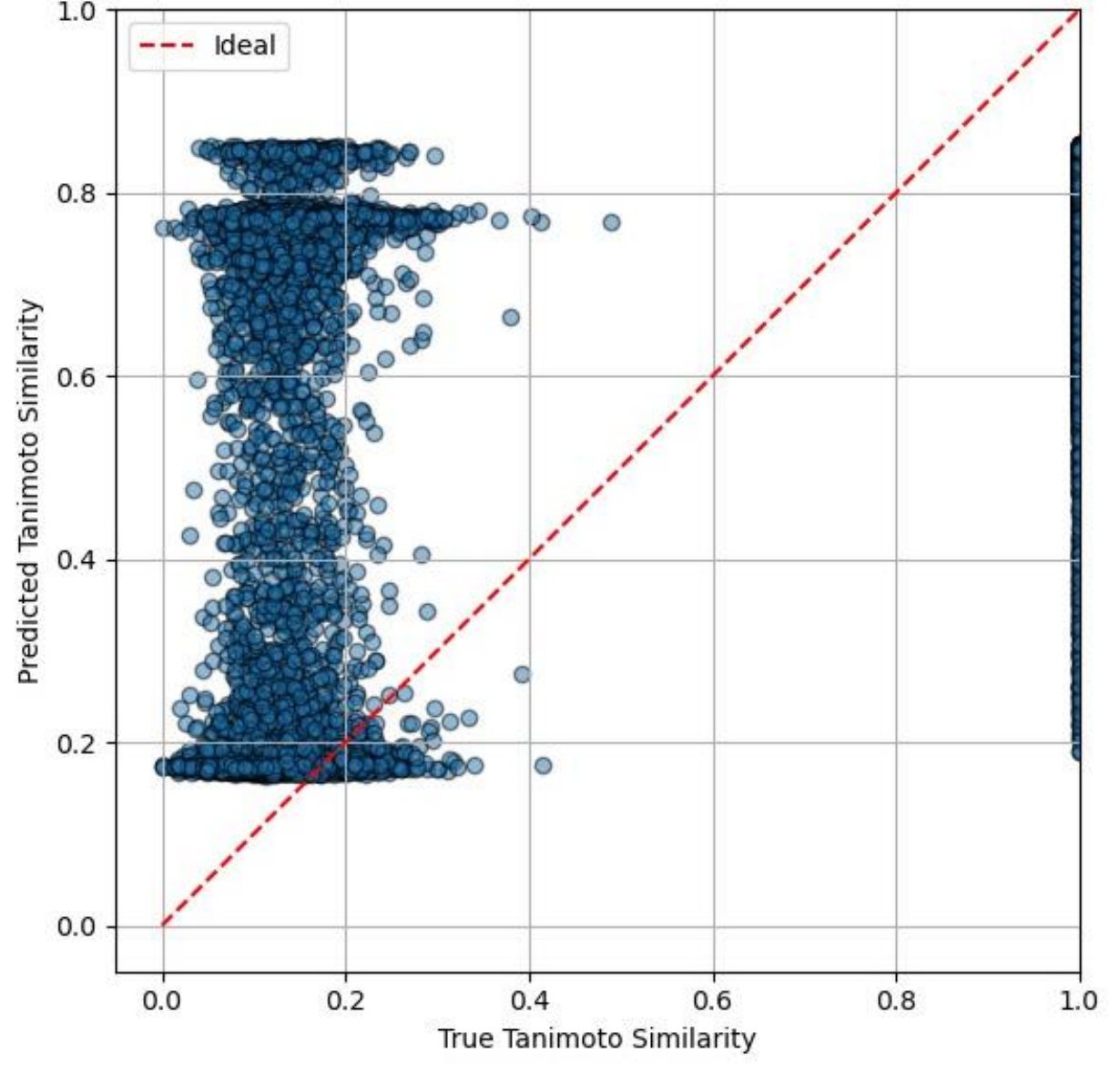}
    \end{subfigure}
    \begin{subfigure}[t]{0.45\textwidth}
        \centering
        \includegraphics[width=\linewidth]{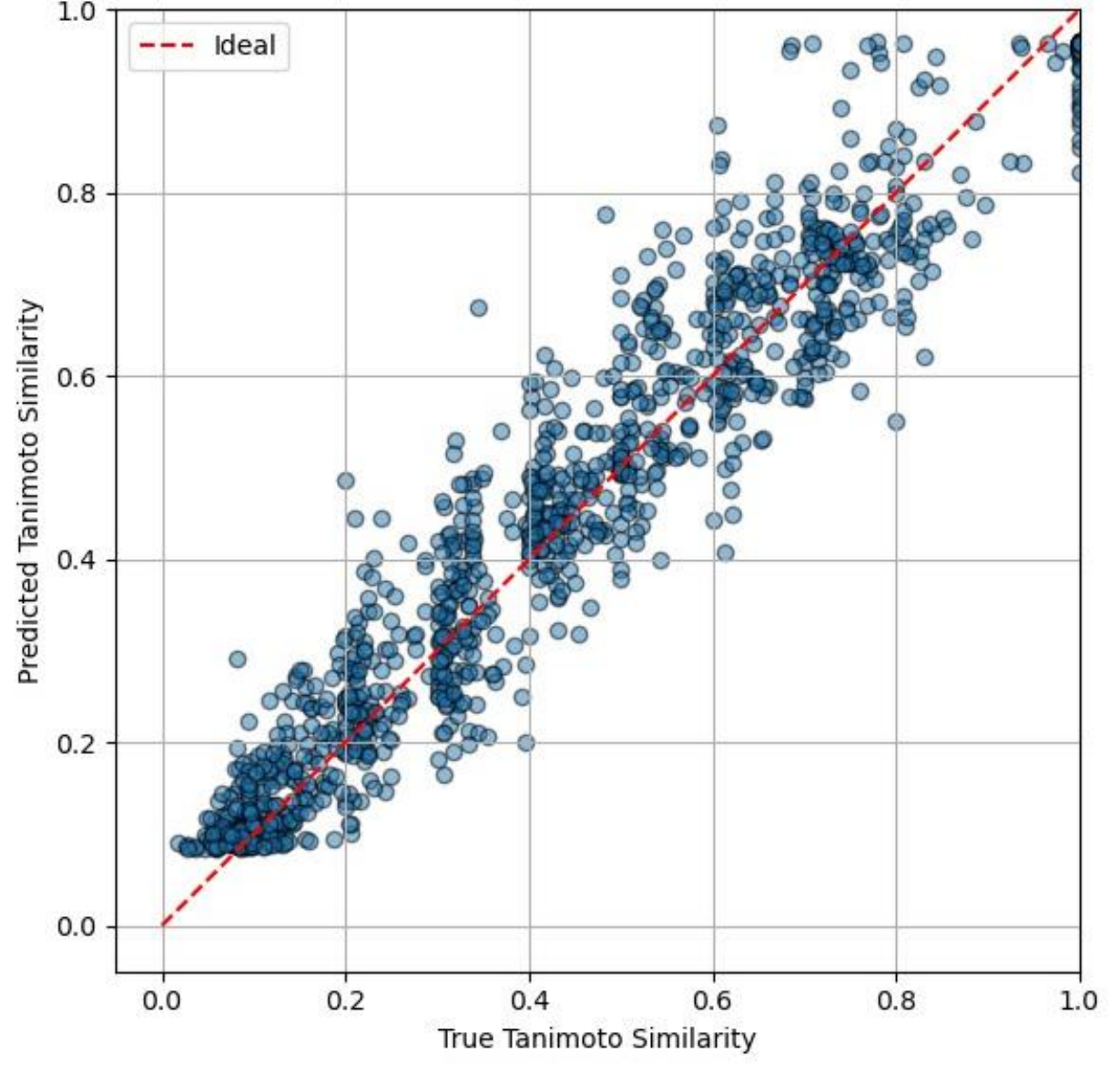}
    \end{subfigure}
    \caption{Left panel: The predicted Tanimoto similarity by $\widehat{f}_2$ trained on raw dataset. Most of the real Tanimoto similarity randomly distributes in $[0.0,0.4] \cup \{1\}$ and the predicted value is severely biased. Right panel: The prediction results of $\widehat{f}_2$ trained on dataset constructed by KNN.}
    \label{fig_tanimoto}
\end{figure}

Let $f_2$ in Section \ref{sec_nearby_sampling} be the function computing the Tanimoto similarity between the source molecule and another molecule. Since $f_2$ is not differentiable \cite{Kaufman_2024_coatildm}, we estimate $f_2$ using a learned regressor $\widehat{f}_2$. Learning the $\widehat{f}_2$ is substantially more challenging than standard regression tasks. The difficulty arises from the sample size of molecular datasets and the imbalanced similarity between random pairs. For the former, computing the similarity between all molecules in  $D^{(1)} \cup D^{(2)}$ requires a computational complexity of $\mathcal{O}((n_1 + n_2)^2)$ where both $n_1$ and $n_2$  exceed 200,000. For the latter, randomly pairing two molecules and computing their similarities result in the vast majority of similarities falling within $[0.0,0.4] \cup \{1\}$. A regressor trained on such randomly sampled similarities is inevitably biased, as shown in Figure \ref{fig_tanimoto}. To address these two issues, we first partition the similarity range [0, 1] into 10 equal intervals and use KNN to identify the 1000 nearest latent vectors for each molecule and compute their corresponding Tanimoto similarities. Once each interval has accumulated 10,000 molecular pairs, we stop constructing the similarity dataset and use it to train a regressor predicting Tanimoto similarity. The prediction results of $\widehat{f}_2$ trained on different datasets are presented in Figure~\ref{fig_tanimoto}.

To perform partial diffusion, we add noise with variance 1 to the latent vector of source molecules and apply a reverse process with guidance scales $\lambda_1 = 230, \lambda_2 = 100$.

\subsection{Diffusion models trained on aggregated vs. single datasets}
One of our core arguments is that, compared to training on a single dataset, leveraging aggregated datasets is expected to yield a better diffusion model. To validate this claim, we compare models trained solely on a single dataset $D^{(1)}$ with those trained on $D^{(1)} \cup D^{(2)}$ under Setting II in Appendix \ref{sec_app_gaussian_mix}. When using only $D^{(1)}$, we assume the access to $\widehat{f}_2$. 

The results can be found in Table \ref{tab_single_aggregate}. Diffusion models trained on $D^{(1)} \cup D^{(2)}$ exhibit consistently smaller $W_2$ distances than those trained on $D^{(1)}$, which supports our claim.

\begin{table}[htbp]
  \centering
  \caption{$W_2$ distances and standard deviations of diffusion models trained on $D^{(1)}$ and $D^{(1)} \cup D^{(2)}$. All experiments are repeated 50 times and evaluated on 1000 samples.}
  \label{tab_single_aggregate}

  \begin{tabular}{lcc}
    \toprule
    \textbf{Method}
      & \textbf{$W_2$ ($\downarrow$) on $D^{(1)}$}
      & \textbf{$W_2$ ($\downarrow$) on $D^{(1)} \cup D^{(2)}$} \\
    \midrule
    DMDG  & 4.506(5.113) & 3.275(3.411) \\
    DMIDG & 6.787(8.211) & 4.688(6.447) \\
    DMHG  & 2.752(3.306) & 2.386(2.923) \\
    DMIHG & 5.947(6.865) & 4.919(6.415) \\
    \bottomrule
  \end{tabular}
\end{table}

\begin{figure}[ht!]
  \centering
  \includegraphics[width=0.4\textwidth]{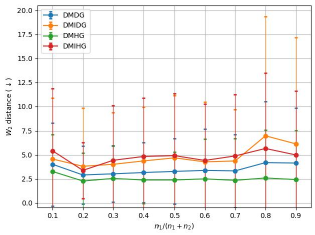}
  \caption{The $W_2$ distances and standard deviations of DMDG, DMIDG, DMHG and DMIHG when $n_1/(n_1+n_2)$ varies.}
  \label{fig_varying_sample_size}
\end{figure}

\subsection{Aggregated datasets with varying sample sizes}
In Section \ref{sec_mol} and Appendix \ref{sec_app_gaussian_mix}, the datasets $D^{(1)}$ and $D^{(2)}$ have equal or at least similar sample sizes. However, we are also interested in the effectiveness of proposed methods when the sample sizes of $D^{(1)}$ and $D^{(2)}$ are imbalanced. To this end, we continue to evaluate different methods with Setting II in Appendix \ref{sec_app_gaussian_mix}. Let $n_1$ and $n_2$ denote the sample sizes of $D^{(1)}$ and $D^{(2)}$, respectively. We fix the total number of samples to $n_1 + n_2 = 20000$, and vary the ratio $n_1 / (n_1+n_2)$ from 0.1 to 0.9. We then evaluate each method by computing the $W_2$ distance between the generated samples and the true conditional distribution $P_{X_0 | C_1, C_2}$. 

The results are shown in Figure~\ref{fig_varying_sample_size}. Our proposed DMDG and DMHG consistently achieve the better performance and exhibit improved stability compared to DMIDG and DMIHG.

\subsection{Applying CG and CFG to flow matching}
\label{sec_app_fm_to_guidance}
JIT adopts flow matching (FM) as its generative framework, which differs from the EDM formulation discussed in Section~\ref{sec_edm} \cite{li_2025_jit,karras_edm,lipman_2023_fm}. As a result, we need to clarify how to add additional guidance into the FM framework. In FM, an ODE, rather than a SDE, is used to define both the noising and reverse processes. Specifically, the ODE is given by:
\begin{equation}
\label{eq_fm_ode}
    \frac{d X_t}{d t} = \epsilon - X_0, \ t \in [0,1], \ \epsilon \sim N(0, I_d),
\end{equation}
which implies that:
\begin{equation*}
\label{eq_fm_interpolation}
    X_t {=} (1 - t) X_0 + t \epsilon.
\end{equation*}
During the generation, we cannot know the target $X_0$. As a result, the right hand side of the ODE in (\ref{eq_fm_ode}) is replaced by $v(X_t, t) := \mathbb{E}[\epsilon-X_0|X_t]$ and the ODE for generation process is:
\begin{equation*}
\label{eq_fm_ode_xt}
    \frac{d X_t}{d t} = v(X_t, t) = \dfrac{1}{t} X_t - \dfrac{1}{t} \mathbb{E}[X_0 | X_t] = \frac{X_t + t \cdot \nabla \log p_t(X_t)}{t - 1},
\end{equation*}
where $\nabla \log p_t(X_t) = -\dfrac{1}{t} \mathbb{E}[\epsilon | X_t] = \dfrac{1-t}{t^2} \mathbb{E}[X_0 | X_t] - \dfrac{1}{t^2} X_t$ \cite{albergo_2023_si}.  Similarly to EDM, the velocity field $v(X_t, t)$ is also intractable, as it depends on the intractable expectation $\mathbb{E}[X_0|X_t]$ or score $\nabla \log p_t(X_t)$. JIT estimates $\mathbb{E}[X_0 | X_t]$, thus $ \nabla \log p_t(X_t)$ and the velocity field $v(X_t, t)$
by minimizing~(\ref{eq_edm_loss}).
As a result, in the FM setting, the unconditional score function can be approximated as:
\begin{equation*}
    \nabla \log p_t(X_t) \approx \dfrac{1-t}{t^2} nn_{\theta}(X_t, t, \varnothing) - \dfrac{1}{t^2} X_t.
\end{equation*}
By replacing the unconditional score function with the corresponding conditional score function, conditional sampling can be achieved. In particular, we can approximate the conditional score function using the DMHG (\ref{eq_hybrid_guidance}):
\begin{align}
\label{eq_score_dmhg_for_fm}
    & \nabla \log p_t(X_t | C_1, C_2)  \notag \\
    & \approx s_{\theta}(X_t,t,C_1,C_2) := \dfrac{1-t}{t^2} \left[ (1- \lambda_1) nn_{\theta}(X_t, t, \varnothing) + \lambda_1 nn_{\theta}(X_t, t, C_1) \right] - \dfrac{1}{t^2} X_t - \lambda_2 \nabla \|C_2-\widehat{f}_2(X_{0|t,C_1}) \|_2^2. 
\end{align}
Therefore, in the case of FM, conditional sampling via DMHG  can be achieved by solving the following ODE from $t = 1$ to $t = t_{\min}$:
\begin{equation*}
     \frac{d X_t}{d t} = v_{\theta}(X_t,t,C_1,C_2) := \frac{X_t + t \cdot s_{\theta}(X_t,t,C_1,C_2)}{t - 1}, \ X_1 \sim N(0,I_d).
\end{equation*}
Replacing $X_{0|t,C_1}$ in (\ref{eq_score_dmhg_for_fm}) by $X_{0|t}$  results in DMIHG.

Note that the Tweedie projection is valid only when the noise $\epsilon$ is independent of $X_0$. 
This implies that flow matching trained with minibatch optimal transport \cite{tong_2024_mini_ot} cannot be used here. 
Fortunately, JIT does not employ minibatch optimal transport and it can be used to DMHG.

\subsection{Results of image inpainting task in Section \ref{sec_inpaint}}
\label{sec_app_inpaint}
We use Learned Perceptual Image Patch Similarity (LPIPS) \cite{Zhang_2018_lpips} to evaluate the quality of generated images, with lower LPIPS indicating better perceptual and semantic similarity to the original image $X_0$. To evaluate the conditional control of $C_1$, we use Top-1 and Top-5 accuracy calculated by $\widehat{f}_1$. The structural similarity index (SSIM) and the peak Signal-to-noise ratio (PSNR) \cite{Alain_2010_psnr} are used to evaluate control over the condition $C_2$, with higher SSIM and PSNR indicating better control. 

\begin{figure*}[ht!]
    \centering
    \includegraphics[width=0.8\linewidth]{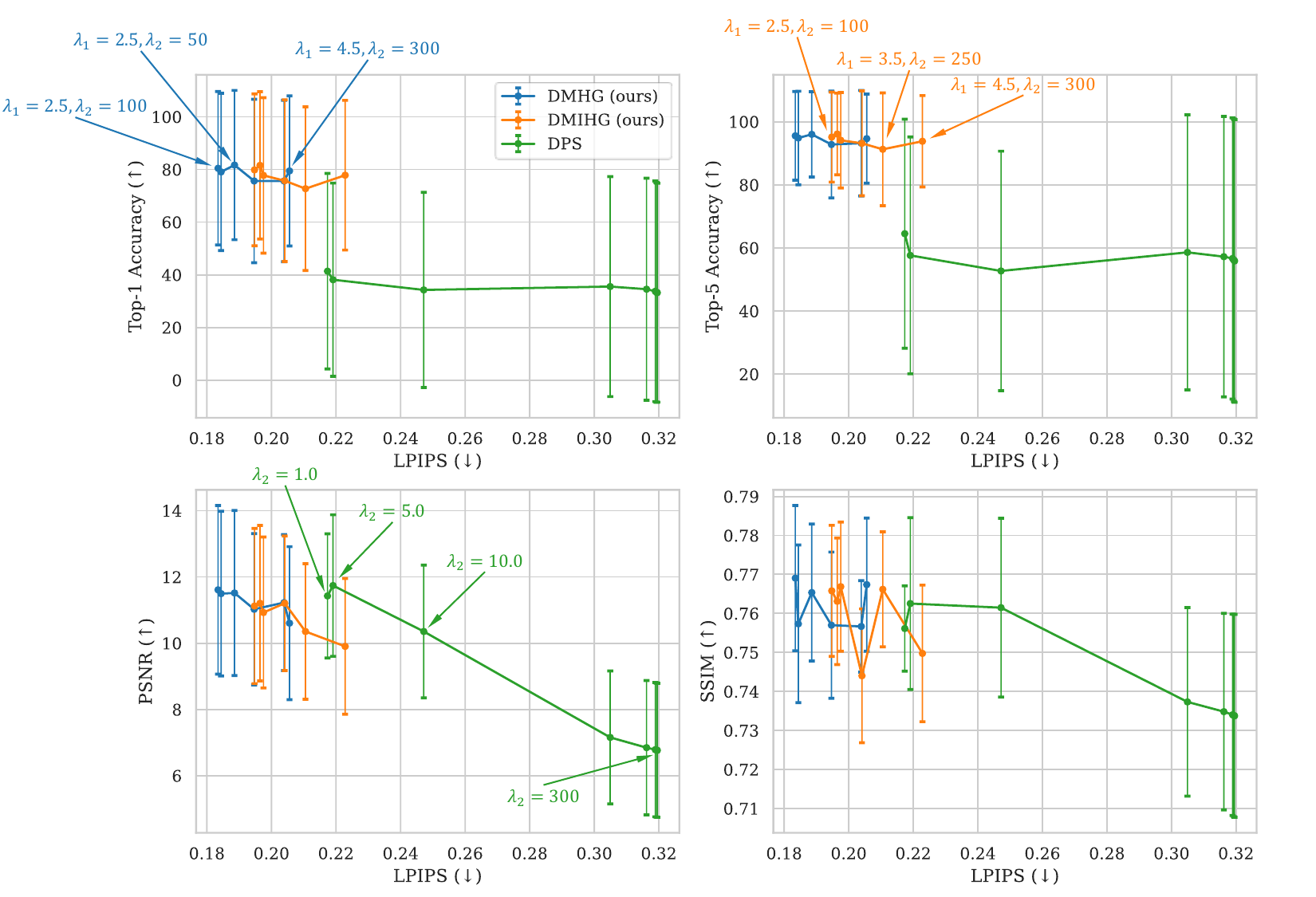}

    \vspace{0.2em}
    \rule{0.8\linewidth}{0.3pt}
    \vspace{0.2em}

    \includegraphics[width=0.8\linewidth]{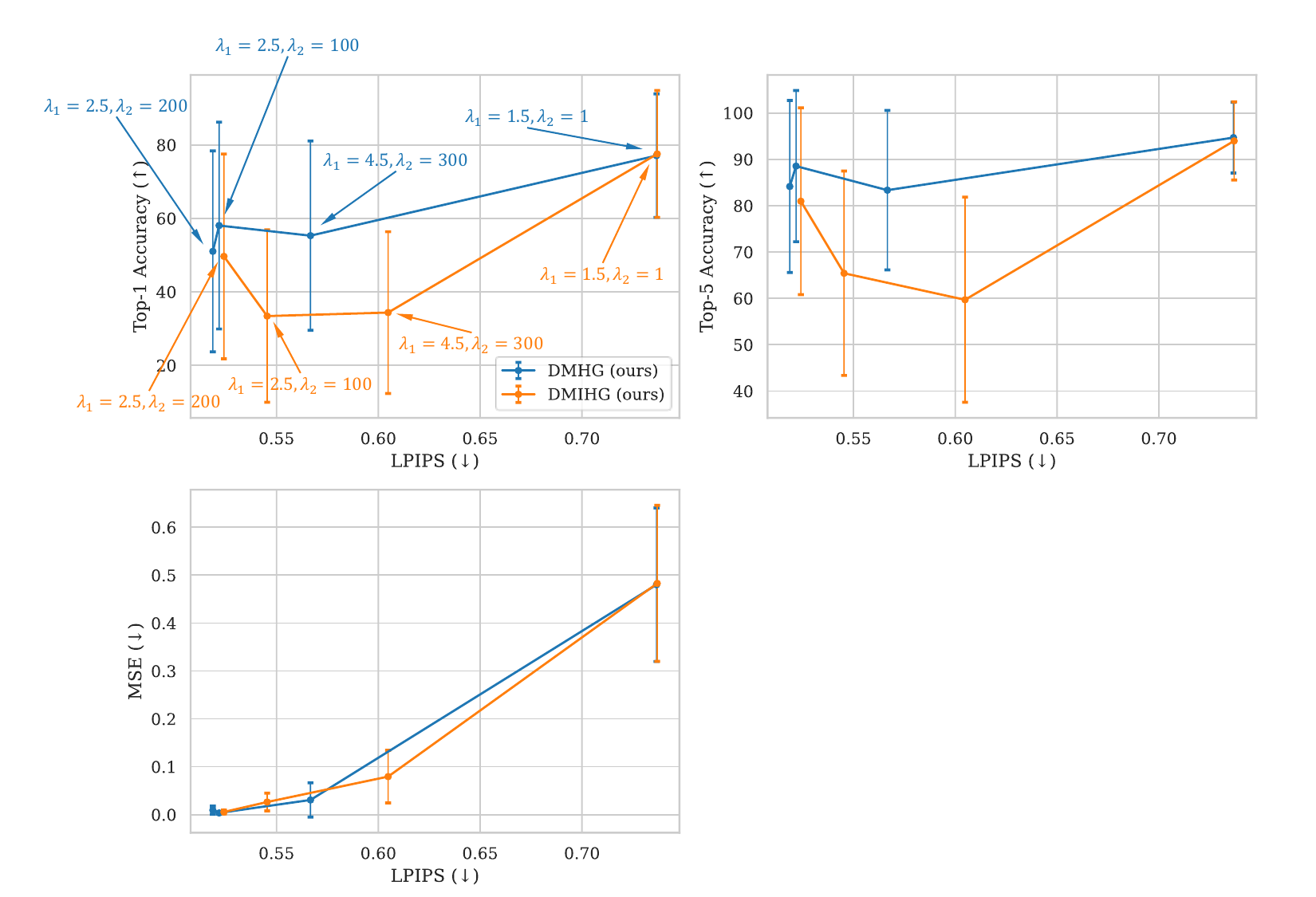}
    \caption{Top panel: results of image inpainting in Section \ref{sec_inpaint}. Bottom panel: results of object editing in Section \ref{sec_obj_edit}. All results are sorted by LPIPS and some of the guidance scales are marked.}
    \label{fig_inpaint_edit_lpips}
\end{figure*}

The inpainting results are shown in the top panel of Figure~\ref{fig_inpaint_edit_lpips}. As can be observed, double guidance methods, namely DMHG and DMIHG, consistently outperform the single-guidance baseline DPS across all metrics, highlighting the substantial benefits of incorporating object information into the inpainting process. Both DMHG and DMIHG achieve similar control over conditions $C_1$ and $C_2$; however, DMHG achieves a noticeably lower LPIPS, indicating higher perceptual quality of the generated images.

\subsection{Results of object editing task in Section \ref{sec_obj_edit}}
\label{sec_app_edit}

When $f_2$ corresponds to a $\times 16$ down-sampling operation, the goal is to preserve the basic background of the image, such as its color tone, while generating an image containing a different object, i.e., the object editing. In this setting, DPS cannot perform object editing. Thus, we only compare DMHG and DMIHG. In addition, randomly selecting a target object leads to a low success rate. Therefore, we choose the target object as the second most likely class predicted by $\widehat{f}_1$.

In object editing, we can still use $\widehat{f}_1$ to measure the model’s control over $C_1$. However, when evaluating control over $C_2$, pixel-wise metrics such as PSNR and structural similarity metrics such as SSIM are no longer effective. Instead, we measure control over $C_2$ using the mean squared error (MSE) between the down-sampled generated images obtained via $f_2$ and the original low-resolution observations $C_2$, and refer to this metric as MSE.

The results of object editing are shown in the bottom panel of Figure~\ref{fig_inpaint_edit_lpips}. As can be observed, DMHG and DMIHG achieve comparable LPIPS and MSE, with DMHG having a slightly lower LPIPS. Despite similar image quality, DMHG exhibits significantly stronger control over $C_1$, indicating a better understanding of the editing objective and more effective generation of target objects.

\subsection{Results of inpainting-based object editing}
\label{sec_app_obj_edit_and_inpaint}

\begin{figure}[htbp]
  \centering
  \includegraphics[width=0.95\textwidth]{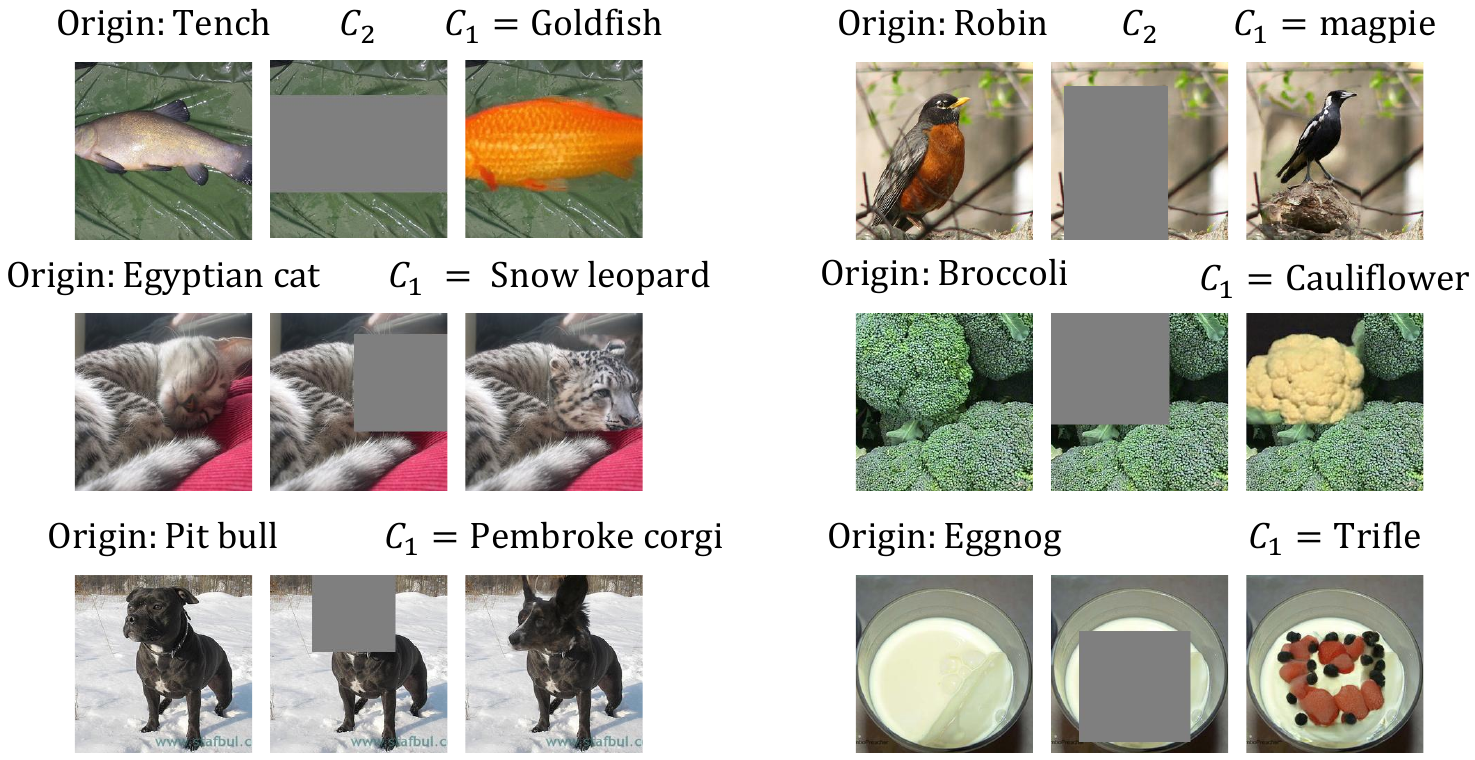}
  \caption{Results of inpainting-based object editing via DMHG. The guidance scales are $\lambda_1 = 2.5$ and $\lambda_2 = 100$}
  \label{fig_inpaint_edit}
\end{figure}

As discussed in Section~\ref{sec_obj_edit}, object editing can also be achieved when $f_2$ corresponds to a masking operation. However, this requires manually masking semantically relevant regions of the image, which makes quantitative evaluation of the generated results infeasible. Therefore, we present representative examples of inpainting-based object editing in Figure~\ref{fig_inpaint_edit}.

\section{Implementation details}
\label{sec_experimental_details}

For molecular generation in Section \ref{sec_mol}, We adopted the same architecture as COATI-LDM \cite{Kaufman_2024_coatildm}. Specifically, the U-Net is built using 2-layer weight-normalized SwiGLU layers, and performs three stages of downsampling followed by three stages of upsampling. Additionally, $C_1$ and $C_2$ were encoded using sine-cosine embeddings. The training, validation, and test sets are split in a ratio of 76\% / 14\% / 10\%, same with \cite{hoogeboom_2022_e3_diffusion}. The probability of masking $C_1$, $p_{\text{non}}$ in (\ref{eq_edm_loss}) was set to $0.5 \cdot n_1/(n_1+n_2)$. We used the AdamW optimizer with a learning rate of 1e-3 and a weight decay of 1e-4. The learning rate was scheduled using cosine annealing with a minimum value of 1e-5. The model was trained for 2400 epochs, which took approximately 20 hours on a single NVIDIA A100. The guidance networks $\widehat{f}_1$ and $\widehat{f}_2$ are variants of the U-Net architecture, with only four downsampling stages and outputting predictions in the final layer. The optimizer is Adam with a learning rate of 1e-3. The maximum epoch was set to 50, and the model with the lowest loss was selected. During CTRL fine-tuning, we used the AdamW optimizer with a learning rate of 5e-4, and set $\gamma$ to 500. The maximum number of training epochs was set to 1000, and the model with the lowest loss was selected. When applying GAIN to the aggregated GEOM and ZINC250k datasets for imputation, we increased the width of the hidden layers and replaced the original fully connected network with a 6-layer residual neural network. The optimizer used was Adam with a learning rate of 1e-3. For Forest Diffusion, due to the large size of the input dataset, we set the duplication factor to 1 and used LightGBM (LGB) as the base model, with all other parameters kept at their default values.

For image generation in Section \ref{sec_image_gen}, we used JIT-H-16 trained on ImageNet at a resolution of $256 \times 256$. We employed ResNet-18 
\cite{Kaiming_2016_resnet}
to evaluate the control of $C_1$. To adapt ResNet-18 to the $3 \times 256 \times 256$ input, we used the AdamW optimizer to fine tune the last layer of ResNet-18, with a learning rate of 1e-4 for the last layer, and 1e-6 for the original parameters. The epoch was set to 4. After fine-tuning, the Top-1 accuracy is 78.03\% and the Top-5 accuracy is 94.07\%.

The network $nn_{\theta}$ in Appendix \ref{sec_app_gaussian_mix} was consistently implemented as a fully connected neural network with 3 hidden layers, each containing 128 neurons and using the SiLU activation function. The guidance networks $\widehat{f}_1$ and $\widehat{f}_2$ shared the same structure but with 2 hidden layers. The training, validation, and test sets are split in a ratio of 87\% / 9\% / 4\%. When training both $nn_{\theta}$ and the $\widehat{f}_1$, $\widehat{f}_2$, we used the Adam optimizer with a learning rate of 1e-3. $nn_{\theta}$ was trained for up to 500 epochs, and the model with the lowest loss was selected. The numbers of training epochs for $\widehat{f}_1$ and $\widehat{f}_2$ were fixed at 50. The network of CTRL \cite{zhao_2025_adding_cond_by_rl} is based on the pre-trained $nn_{\theta}$ and is fine-tuned using the ControlNet \cite{zhang_2023_control_net}. The optimizer of fine-tuning remains Adam, but the learning rate is set to 1e-4. Training is performed for up to 150 epochs, and the model with the lowest loss is selected. To compute the integral in (\ref{eq_rl_funetune_loss}), we first perform full sampling using EDM, with the number of sampling steps drawn from $\mathcal{U}\{18,19,\hdots,128\}$. In the experiments under Setting I, the value of $\gamma$ is set to 180, while in Setting II, it is set to 1.5. When using GAIN \cite{Yoon_2018_gain} and Forest Diffusion \cite{Jolicoeur_2024_forestdiffusion} for imputation, we adopted the default hyperparameters from their Github repositories, except that the hidden layers of the GAIN network were doubled in width to enhance its fitting capacity.  

Unless otherwise specified, all diffusion models discussed in this paper refer to EDM and the sampling steps of the reverse process are set to 18. The sampling steps of JIT model used in Section \ref{sec_image_gen} is 50.

\section{Computational Efficiency}

\begin{figure}[htbp]
  \centering
  \includegraphics[width=0.9\textwidth]{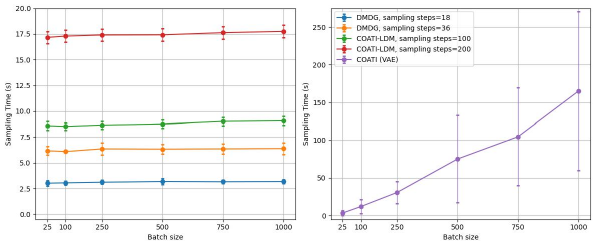}

  \vspace{0.2em}
    \rule{0.9\linewidth}{0.3pt}
    \vspace{1.6em}

    \includegraphics[width=0.5\linewidth]{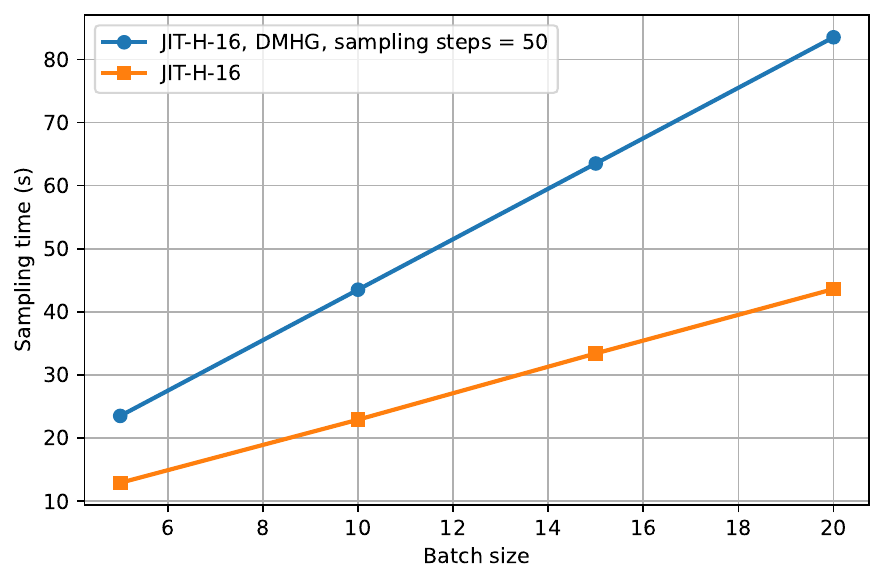}

  \caption{Sampling time comparison of De novo drug design (upper panel) and image inpainting (bottom panel) under varying batch sizes. The upper-left panel shows the sampling time of EDM with different sampling steps. The upper-right panel displays the decoding time of COATI across the same batch sizes. The bottom panel shows the sampling time of JIT and DMHG with different batch sizes.}
  \label{fig_sampling_speed}
\end{figure}

Thanks to the simplicity of the EDM framework, we are able to obtain high-quality samples using only a small number of sampling steps \cite{karras_edm}. Unless otherwise specified, the number of sampling steps is set to 18, which brings significant advantages in computational efficiency. In Figure \ref{fig_sampling_speed}, we report the sampling speed of our model and COATI-LDM under varying batch sizes, along with the decoding time of COATI. For visualization purposes, we set the number of sampling steps of COATI-LDM to 100 and 200, while its default sampling steps is actually 1000. All measurements were conducted on an NVIDIA A100 GPU. As observed, when the sampling step is set to 18, DMDG achieves the fastest sampling speed, and the runtime remains relatively stable across different batch sizes. In contrast, decoding latent vectors using COATI is highly sensitive to the batch size. Larger batches significantly slow down the decoding process. This suggests that a smaller batch size should be preferred when decoding with COATI. Furthermore, our experiments in Section \ref{sec_nearby_sampling} demonstrate that DMDG maintains a high nearby sampling success rate even under limited oracle calls, which highlights one of its advantages.

In Figure \ref{fig_sampling_speed}, we also report the sampling speed of image inpainting via DMHG and JIT-H-16 in Section \ref{sec_inpaint}. It can be observed that sampling with DMHG is approximately twice as fast as sampling with JIT alone, since DMHG requires computing gradients through the entire JIT model and ResNet-18.

\section{Proof of theorems}
\label{sec_app_proof}
\subsection{Proof of Theorem \ref{thm_square_of_densities}}
We only provide the upper bound of $| p(C_2|X_t,C_1) - p(C_2|X_{0|t,C_1}) |^2$ since the other one is similar.
\begin{align}
\label{eq_pf_of_thm1}
    &| p(C_2|X_t,C_1) - p(C_2|X_{0|t,C_1}) |^2 \notag \\
    & \leq | p(C_2|X_t,C_1) - p(C_2|\mathbb{E}(X_0|X_t,C_1)) |^2 & (I) \notag \\
    & \ \ \ \ + | p(C_2|\mathbb{E}(X_0|X_t,C_1)) - p(C_2|X_{0|t,C_1})|^2. & (II)
\end{align}
Then, we bound $(I)$ and $(II)$ respectively. For $(I)$, we have:
\begin{align}
\label{eq_leq_I}
    & | p(C_2|X_t,C_1) - p(C_2|\mathbb{E}(X_0|X_t,C_1)) |^2 \notag \\
    & = | \mathbb{E}_{X_0 \sim P_{X_0|X_t,C_1}} \left[ p(C_2|X_0) - p(C_2|\mathbb{E} \left[ X_0|X_t,C_1 \right]) \right] |^2 \notag \\
    & \leq \mathbb{E}_{X_0 \sim P_{X_0|X_t,C_1}} | p(C_2|X_0) - p(C_2
    |\mathbb{E} \left[ X_0|X_t,C_1 \right]) |^2.
\end{align}
Let $\phi( \mu, \sigma^2)$ be the density function of a normal distribution associated with the variable $C_2$, having mean $\mu$ and covariance $\sigma^2 I_{p-k}$. Following \cite{chung_diffusion_inverse}, we have the following lemma:
\begin{lemma}
\label{thm_gauss_with_equal_var_and_different_mu}
    $\forall$ $\mu_1,\mu_2$ in $\mathbb{R}^{p-k}$,
    \begin{align}
        |\phi(\mu_1, \sigma_2^2) - \phi(\mu_2, \sigma_2^2)  |^2 \leq \dfrac{p-k}{\sqrt{2 \pi} \sigma_{2}} \exp \left( - \dfrac{1}{2 \sigma_{2}^2} \right) \| \mu_1 - \mu_2 \|_2^2.
    \end{align}
\end{lemma}
Then, $| p(C_2|X_0) - p(C_2 |\mathbb{E} \left[ X_0|X_t,C_1 \right]) |^2 $ can be bound by:
\begin{align}
\label{eq_lemma}
    & | p(C_2|X_0) - p(C_2 |\mathbb{E} \left[ X_0|X_t,C_1 \right]) |^2 \notag \\
    & = |\phi(f_2(X_0), \sigma_2^2) - \phi(f_2(\mathbb{E} \left[ X_0|X_t,C_1 \right]), \sigma_2^2)  |^2 \notag \\
    & \leq \dfrac{p-k}{\sqrt{2 \pi} \sigma_{2}} \exp \left( - \dfrac{1}{2 \sigma_{2}^2} \right) \| f_2(X_0) - f_2(\mathbb{E} \left[ X_0|X_t,C_1 \right]) \|_2^2 \notag \\
    & \leq \dfrac{p-k}{\sqrt{2 \pi} \sigma_{2}} \exp \left( - \dfrac{1}{2 \sigma_{2}^2} \right) \cdot \max_{z \in \mathbb{R}^{d}} \| \nabla_{z} f_2(z) \|^2_2  \cdot \| X_0-  \mathbb{E} \left[ X_0|X_t,C_1 \right] \|_2^2.
\end{align}
Then, by plugging (\ref{eq_lemma}) into (\ref{eq_leq_I}), it can be derived that the upper bound of $(I)$ is:
\begin{align}
\label{eq_I_ineq}
    & |p(C_2|X_t,C_1) - p(C_2|\mathbb{E} \left[ X_0|X_t,C_1 \right]) |^2 \notag \\
    & \leq \dfrac{p-k}{\sqrt{2 \pi} \sigma_{2}} \exp \left( - \dfrac{1}{2 \sigma_{2}^2} \right) \cdot \max_{z \in \mathbb{R}^d} \| \nabla_{z} f_2(z) \|_2^2  \cdot \mathbb{E}_{X_0 \sim P_{X_0|X_t,C_1}} \| X_0 - \mathbb{E} \left[ X_0|X_t,C_1 \right] \|_2^2.
\end{align}
It is worthy noting that $\max_{z \in \mathbb{R}^d} \| \nabla_{z} f_2(z) \|_2^2$ and $\mathbb{E}_{X_0 \sim P_{X_0|X_t,C_1}} \| X_0 - \mathbb{E} \left[ X_0|X_t,C_1 \right] \|_2^2$ changes with the $f_2$ and $P_{X_0}$, but in most real cases, they are bounded \cite{chung_diffusion_inverse}.

Next, we will demonstrate how to derive an upper bound for $(II)$, in which $p(C_2|\mathbb{E} \left[ X_0|X_t,C_1 \right])=\phi(f_2(\mathbb{E} \left[ X_0|X_t,C_1 \right]),\sigma_2^2 )$ and $p(C_2|X_{0|t,C_1}) = \phi(\widehat{f}_2(X_{0|t,C_1}), 1/2\lambda_2)$. We have:
\begin{align}
\label{eq_II_ineq}
    & | p(C_2|\mathbb{E} \left[ X_0|X_t,C_1 \right]) - p(C_2|X_{0|t,C_1}) |^2 \notag \\
    & \leq | \phi(f_2(\mathbb{E} \left[ X_0|X_t,C_1 \right]),\sigma_2^2 ) - \phi(\widehat{f}_2(X_{0|t,C_1}),\sigma_2^2 ) |^2 + | \phi(\widehat{f}_2(X_{0|t,C_1}) ,\sigma_2^2 ) - \phi( \widehat{f}_2(X_{0|t,C_1}) ,1/2\lambda_2 ) |^2 \notag \\
    & \overset{(a)}{\leq} | \phi(f_2(\mathbb{E} \left[ X_0|X_t,C_1 \right]),\sigma_2^2 ) - \phi(\widehat{f}_2(X_{0|t,C_1}),\sigma_2^2 ) |^2 + U_{2} (\sigma^2_2 -1/2\lambda_2)^2 \notag \\
    & \overset{(b)}{\leq} \dfrac{p-k}{\sqrt{2 \pi} \sigma_{2}} \exp \left( - \dfrac{1}{2 \sigma_{2}^2} \right) \|  f_2(\mathbb{E} \left[ X_0|X_t,C_1 \right]) - \widehat{f}_2(X_{0|t,C_1}) \|_2^2  + U_{2} (\sigma^2_2 -1/2\lambda_2)^2. 
\end{align}
For $(a)$, we use mean value theorem and $| \phi'_{\sigma}(\mu,\sigma^2) |^2$ has a upper bound $U_{2}$ when $\sigma^2 \in \left[ \min(\sigma^2_2, 1/2\lambda_2) , \max(\sigma^2_2, 1/2\lambda_2) \right]$, since $ \phi'_{\sigma}(\cdot, \sigma^2)$ is continuous in the interval. For $(b)$, we used Lemma \ref{thm_gauss_with_equal_var_and_different_mu}. Then, we further bound $\|  f_2(\mathbb{E} \left[ X_0|X_t,C_1 \right]) - \widehat{f}_2(X_{0|t,C_1}) \|_2^2$.
\begin{align}
\label{eq_2_funciton_2norm}
    & \|  f_2(\mathbb{E} \left[ X_0|X_t,C_1 \right]) - \widehat{f}_2(X_{0|t,C_1}) \|_2^2 \notag \\
    & \leq \| f_2(\mathbb{E} \left[ X_0|X_t,C_1 \right]) - \widehat{f}_2(\mathbb{E} \left[ X_0|X_t,C_1 \right]) \|_2^2 + \|  \widehat{f}_2(\mathbb{E} \left[ X_0|X_t,C_1 \right]) - \widehat{f}_2(X_{0|t,C_1})  \|_2^2 \notag \\
    & \overset{(c)}{\leq} \| f_2(\mathbb{E} \left[ X_0|X_t,C_1 \right]) -  \mathbb{E}_{X_0 \sim P_{X_0|X_t,C_1}} \left[ f_2(X_0) \right]\|_2^2 \notag \\
    & \ \ \ \ + \| \mathbb{E}_{X_0 \sim P_{X_0|X_t,C_1}} \left[ f_2(X_0) \right] - \mathbb{E}_{X_0 \sim P_{X_0|X_t,C_1}} [ \widehat{f}_2(X_0) ]   \|_2^2 \notag \\
    & \ \ \ \ + \| \mathbb{E}_{X_0 \sim P_{X_0|X_t,C_1}} [ \widehat{f}_2(X_0) ] - \widehat{f}_2(\mathbb{E} \left[ X_0 | X_t,C_1 \right]) \|_2^2 \notag \\
    & \ \ \ \ + \max_{z \in \mathbb{R}} \| \nabla_{z} \widehat{f}(z) \|_2^2 \cdot \| \mathbb{E} \left[ X_0|X_t,C_1 \right] - X_{0|t,C_1} \|_2^2 \notag \\
    & \overset{(d)}{\leq} \mathbb{E}_{X_0 \sim P_{X_0|X_t,C_1}} \| f_2(\mathbb{E} \left[ X_0|X_t,C_1 \right]) - f_2(X_0) \|_2^2 \notag \\
    & \ \ \ \ + \mathbb{E}_{X_0 \sim P_{X_0|X_t,C_1}} \| f_2(X_0) - \widehat{f}_2(X_0) \|_2^2 \notag \\
    & \ \ \ \ + \mathbb{E}_{X_0 \sim P_{X_0|X_t,C_1}} \| \widehat{f}_2(X_0) - \widehat{f}_2(\mathbb{E} \left[ X_0|X_t,C_1 \right]) \|_2^2 \notag \\
    & \ \ \ \ + \max_{z \in \mathbb{R}} \| \nabla_{z} \widehat{f}(z) \|_2^2 \cdot \| \mathbb{E} \left[ X_0|X_t,C_1 \right] - X_{0|t,C_1} \|_2^2 \notag \\
    & \overset{(e)}{\leq} \max_{z \in \mathbb{R}} \| \nabla_{z} f(z) \|_2^2 \cdot \mathbb{E}_{X_0 \sim P_{X_0|X_t,C_1}} \| \mathbb{E} \left[ X_0|X_t,C_1 \right] - X_0 \|_2^2 \notag \\
    & \ \ \ \ + \mathbb{E}_{X_0 \sim P_{X_0|X_t,C_1}} \| f_2(X_0) - \widehat{f}_2(X_0) \|_2^2 \notag \\
    & \ \ \ \ + \max_{z \in \mathbb{R}} \| \nabla_{z} \widehat{f}(z) \|_2^2 \cdot \mathbb{E}_{X_0 \sim P_{X_0|X_t,C_1}} \| X_0 - \mathbb{E} \left[ X_0|X_t,C_1 \right] \|_2^2 \notag \\
    & \ \ \ \ + \max_{z \in \mathbb{R}} \| \nabla_{z} \widehat{f}(z) \|_2^2 \cdot \| \mathbb{E} \left[ X_0|X_t,C_1 \right] - X_{0|t,C_1} \|_2^2.
\end{align}
In $(c), (e)$, we used the mean value theorem and in $(d)$, we used the Jensen inequality same in (\ref{eq_leq_I}).

We want a more concise upper bound. So, let us dive deeper into $\max_{z \in \mathbb{R}^d} \| \nabla_{z} \widehat{f}_2(z) \|_2^2$ and the conditional variance $\mathbb{E}_{X_0 \sim p(X_0|X_t)} \| X_0 - \mathbb{E} \left[ X_0|X_t \right] \|_2^2$. As mentioned in Section \ref{sec_dmdg}, $\widehat{f}_1$ and $\widehat{f}_2$ are ReLU networks. In Lemma \ref{thm_up_bound_of_grad}, we prove the upper bound of $\max_{z \in \mathbb{R}^d} \| \nabla_{z} \widehat{f}_2(z) \|_2^2$. The proof of the upper bound of $\max_{z \in \mathbb{R}^d} \| \nabla_{z} \widehat{f}_1(z) \|_2^2$ follows the same way. 

Combining (\ref{eq_pf_of_thm1}), (\ref{eq_I_ineq}), (\ref{eq_II_ineq}), (\ref{eq_2_funciton_2norm}) and Lemma \ref{thm_up_bound_of_grad}, we can finally get the error bound:
\begin{align}
\label{eq_bound_of_thm1}
    & | p(C_2|X_t,C_1) - p(C_2|X_{0|t,C_1}) |^2 \notag \\
    & \leq \dfrac{p-k}{\sqrt{2 \pi} \sigma_{2}} \exp \left( - \dfrac{1}{2 \sigma_{2}^2} \right) \Bigl\{ (2 \max_{z \in \mathbb{R}^d} \| \nabla_{z} f_2(z) \|_2^2 + M^L) \notag \\
    & \ \ \ \ \cdot \mathbb{E}_{X_0 \sim P_{X_0|X_t,C_1}} \| X_0 - \mathbb{E} \left[ X_0|X_t,C_1 \right] \|_2^2  \notag \\
    & \ \ \ \ + M^L \| \mathbb{E}\left[ X_0|X_t,C_1 \right] -X_{0|t,C_1} \|_2^2 \notag \\
    & \ \ \ \ + \mathbb{E}_{X_0 \sim P_{X_0|X_t,C_1}} \| f_2(X_0)-\widehat{f}_2(X_0) \|_2^2 \Bigr\} \notag \\
    & \ \ \ \ + U_{2} (\sigma^2_2-1/\lambda_2)^2,
\end{align}
where $M$ and $L$ is the upper bound of the $\ell_2$-norm of the weight matrix, and the number of layers of $\widehat{f}_2$. 

\begin{lemma}
\label{thm_up_bound_of_grad}
    Assume $\widehat{f}_2$ is a ReLU network having $L$ layers and the hidden dimension of each layer is $H$. Let $W_{l}, b_{l}, \ l=1,\cdots,L$ be the weights and biases of $\widehat{f}_2$. Let $\max_{l \in \{ 1,\cdots L \} } \| W_{l} \|_2^2 \leq M$, then we have $\max_{z \in \mathbb{R}^d} \| \nabla_{z} \widehat{f}_2(z) \|_2^2 \leq M^L $.

    Proof: Let $\rho$ be the ReLU function and assume the structure of ReLU network is :
    \begin{align}
        h_{1} &= W_{1} \, z + b_{1} , \notag \\
        h_{2} & = W_{2} \, \rho (h_{1}) + b_{2} , \notag \\
        &\quad \vdots \notag \\
        h_{L} &=  W_{L} \, \rho (h_{L-1}) + b_{L} ,
    \end{align}
    where $z \in \mathbb{R}^d$ is the input, $h_{l} \in \mathbb{R}^H,l=1, \cdots, L-1$ are hidden layers and $h_L \in \mathbb{R}^k$ is the output layer. Then, the gradient of $\widehat{f}_2(z)$ to $z$ is:
    \begin{equation}
        \nabla_{z} \widehat{f}_2(z) = 
        \dfrac{\partial h_{L}}{\partial \rho (h_{L-1})} 
        \dfrac{\partial \rho (h_{L-1})}{\partial h_{L-1}} 
        \cdots
        \dfrac{\partial h_{2}}{\partial \rho (h_{1})} 
        \dfrac{\partial \rho (h_{1})}{\partial h_{1}} \cdot 
        \dfrac{\partial h_{1}}{\partial z}.
    \end{equation}
    It is obvious that $\dfrac{\partial h_{l}}{\partial \rho (h_{l-1})} =W_{l}, \; l=2, \cdots L$ and $\dfrac{\partial h_{1}}{\partial z} = W_{1}$. The partial derivative of the ReLU function $\rho (h_{l})$ with respect to its input $h_{l}$ is:
    \begin{equation*}
        \dfrac{\partial \rho (h_{l})}{\partial h_{l}} = \begin{pmatrix}
            \mathbb{I} \{ h_{l,1} \geq 0 \} & 0 & \cdots & 0 \\
            0 & \mathbb{I} \{ h_{l,2} \geq 0 \} & \cdots & 0 \\
            \vdots & \vdots & \ddots  & \vdots \\
            0 & 0 & \cdots & \mathbb{I} \{ h_{l,H} \geq 0 \}
        \end{pmatrix}
    \end{equation*}
    for $l$ from 1 to $L-1$. And $\mathbb{I} \{ \cdot \}$ is the indicator function and $h_{l,h}$ is the $h-$component of the vector $h_{l}$ for $h = 1,\cdots,H$. So, $ \| \partial \rho (h_{l}) / \partial h_{l} \|_2^2 \leq 1$ for $l=1, \cdots, L-1$. Then, we can get the upper bound of the $\ell_2$-norm of the gradient of $\widehat{f}_1$ to the input $z$:
    \begin{equation}
        \| \nabla_{z} \widehat{f}_2 (z) \|_2^2 \leq \prod_{l=2}^L \| \dfrac{\partial h_{l}}{\partial \rho (h_{l-1})} \|_2^2 \cdot \| \dfrac{\partial h_{1}}{\partial z} \|_2^2 \leq M^L.
    \end{equation}
    
\end{lemma}

\subsection{Bounding the risks of $\widehat{f}_1$ and $\widehat{f}_2$ under distribution shift}
\label{sec_app_th_bound_risk}

In our problem setting, the two regressors $\widehat{f}_1$ and $\widehat{f}_2$ have to be trained on different datasets, $D^{(1)}$ and $D^{(2)}$, respectively, as illustrated by Table \ref{table_block_wise_missing}. 
Due to differences in data collection, a distribution shift between $X_0^{(1)}$ and $X_0^{(2)}$ is generally inevitable, especially when the sample sizes are limited. 
Such distribution shift influences the prediction errors of $\widehat{f}_i,i=1,2$ in Theorem \ref{thm_square_of_densities}. Let $X_0^{(1)} \sim P^{(1)}_{X_0}$ and $X_0^{(2)} \sim P^{(2)}_{X_0}$.  Define $\mathcal{R}_{i}^{(j)}(\widehat{f}_i)=\mathbb{E}_{(X_0,C_i) \sim P^{(j)}_{X_0} \cdot P_{C_i|X_0}}\left[ \ell (\widehat{f}_i(X_0),C_i) \right], (i,j=1,2)$ as the risks with loss function $\ell (\cdot,\cdot)$, where $P_{C_1|X_0}$ and $P_{C_2|X_0}$ are the normal distributions defined in (\ref{eq_function_map}). Each $\widehat{f}_i$ is trained to minimize the risk $\mathcal{R}_i^{(i)}(\widehat{f}_i)$ under its own training distribution. It is crucial that both $\widehat{f}_1$ and $\widehat{f}_2$ generalize well to the alternative distribution. Such generalization is essential for ensuring the effectiveness of the reverse process. 
Theorem \ref{thm_risk_under_shift} reveals that the risks under the shifted distribution also have upper bounds under some conditions.

\begin{theorem}
\label{thm_risk_under_shift}
    Let $p^{(1)}(X_0)$ and $p^{(2)}(X_0)$ be the densities of $P^{(1)}_{X_0}$ and $P^{(2)}_{X_0}$ respectively. Assume the two densities have same support, and both of them have a positive lower bound $\underline{p}>0$. Let $\sup_{X_0' \in \mathbb{R}^d}| p^{(1)}(X_0')-p^{(2)}(X_0') | \leq \delta$. The risks under shifted distribution, denoted as  $\mathcal{R}_1^{(\textcolor{red}{2})}(\widehat{f}_1)$ and $\mathcal{R}_2^{(\textcolor{red}{1})}(\widehat{f}_2)$ have upper bounds:
    \begin{align}
        \mathcal{R}_1^{(\textcolor{red}{2})}(\widehat{f}_1)  &\leq \mathcal{R}_1^{(\textcolor{red}{1})}(\widehat{f}_1) \cdot \exp({\delta}/{\underline{p}}), \notag \\
        \mathcal{R}_2^{(\textcolor{red}{1})}(\widehat{f}_2)  &\leq \mathcal{R}_2^{(\textcolor{red}{2})}(\widehat{f}_2) \cdot \exp({\delta}/{\underline{p}}).
    \end{align}
\end{theorem} \noindent
\textit{proof: 
\begin{align}
    & \mathcal{R}_1^{(\textcolor{red}{2})}(\widehat{f}_1) \notag \\
    & = \mathbb{E}_{(X_0,C_1) \sim P^{(2)}_{X_0} \cdot P_{C_1|X_0}} \left[ \ell (f_1(X_0), C_1)  \right] \notag \\
    & = \int \ell (f_1(X_0), C_1) p^{(\textcolor{red}{2})}(X_0)p(C_1|X_0) d X_0 d C_1 \notag \\
    & =  \int \ell (f_1(X_0), C_1) \dfrac{p^{(\textcolor{red}{2})}(X_0)}{p^{(\textcolor{red}{1})}(X_0)} p(C_1|X_0) p^{(\textcolor{red}{1})}(X_0) d X_0 d C_1 \notag \\
    & = \int \ell (f_1(X_0), C_1)  p(C_1|X_0) p^{(\textcolor{red}{1})}(X_0) d X_0 d C_1  \cdot \sup_{X_0' \in \mathbb{R}^d}  \dfrac{p^{(\textcolor{red}{2})}(X_0')}{p^{(\textcolor{red}{1})}(X_0')} \notag \\
    & =\mathcal{R}_1^{(\textcolor{red}{1})}(\widehat{f}_1) \cdot \sup_{X_0' \in \mathbb{R}^d}  \dfrac{p^{(\textcolor{red}{2})}(X_0')}{p^{(\textcolor{red}{1})}(X_0')} \notag \\
    & \leq \mathcal{R}_1^{(\textcolor{red}{1})}(\widehat{f}_1) \cdot  \exp \left[ \sup_{X_0' \in \mathbb{R}^d} \ln \left(  \dfrac{p^{(\textcolor{red}{2})}(X_0')}{p^{(\textcolor{red}{1})}(X_0')} \right) \right] \notag \\
    & \overset{(f)}{\leq} \mathcal{R}_1^{(\textcolor{red}{1})}(\widehat{f}_1) \cdot \exp \left[ \sup_{X_0' \in \mathbb{R}^d} \dfrac{p^{(\textcolor{red}{2})}(X_0') - p^{(\textcolor{red}{1})}(X_0')}{p^{(\textcolor{red}{1})}(X_0')} \right] \notag \\
    & \leq \mathcal{R}_1^{(\textcolor{red}{1})}(\widehat{f}_1) \cdot \exp (\delta/\underline{p}),
\end{align}
where $(f)$ is derived the inequality $\ln x \leq x-1$ for all $x > 0$. The proof of $\mathcal{R}_2^{(\textcolor{red}{1})}(\widehat{f}_2) \leq \mathcal{R}_2^{(\textcolor{red}{2})}(\widehat{f}_2) \cdot \exp (\delta/\underline{p})$ follows the same way.
}
\begin{remark}
    Theorem \ref{thm_risk_under_shift} means when $\mathcal{R}_1^{(1)}(\widehat{f}_1)$ and $\mathcal{R}_2^{(2)}(\widehat{f}_2)$ are small enough, then their performance on alternative datasets should also be good.
\end{remark}

\section{Selections of Generated Molecules and Images}
\label{sec_selected_mols}
\begin{figure*}[ht!]
  \centering
  \includegraphics[width=0.75\linewidth]{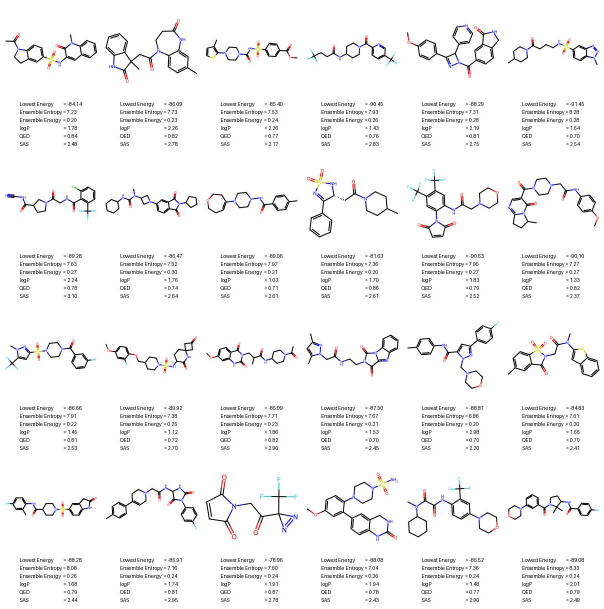}
  \caption{Selected molecules generated by DMDG that satisfy Tasks~1 and~2 (\ref{eq_molecular_task_12}). The guidance scale is $\lambda_1=63,\ \lambda_2=21$.}
  \label{fig_generated_mols_task12}
\end{figure*}

\begin{figure*}[ht!]
  \centering
  \includegraphics[width=0.75\linewidth]{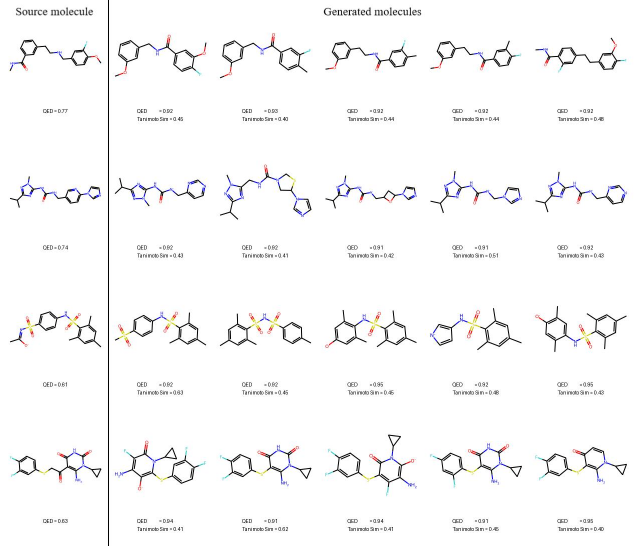}
  \caption{Selected molecules successfully generated by partial DMDG. The guidance scale is $\lambda_1=230,\ \lambda_2=100$.}
  \label{fig_partial_diff_all}
\end{figure*}

\begin{figure*}[ht!]
    \centering
    \includegraphics[width=0.75\linewidth]{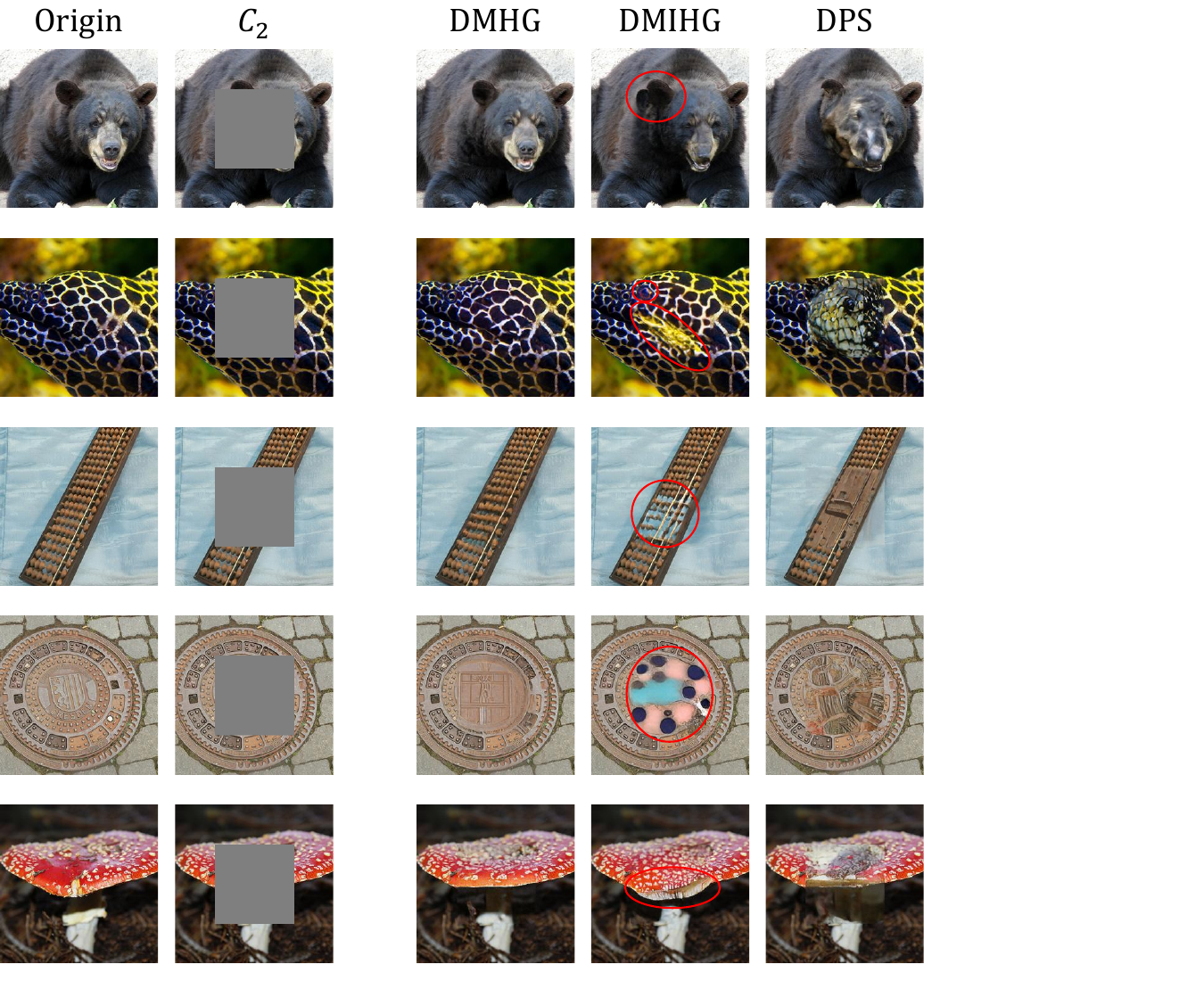}

    \rule{0.8\linewidth}{0.3pt}

    \includegraphics[width=0.75\linewidth]{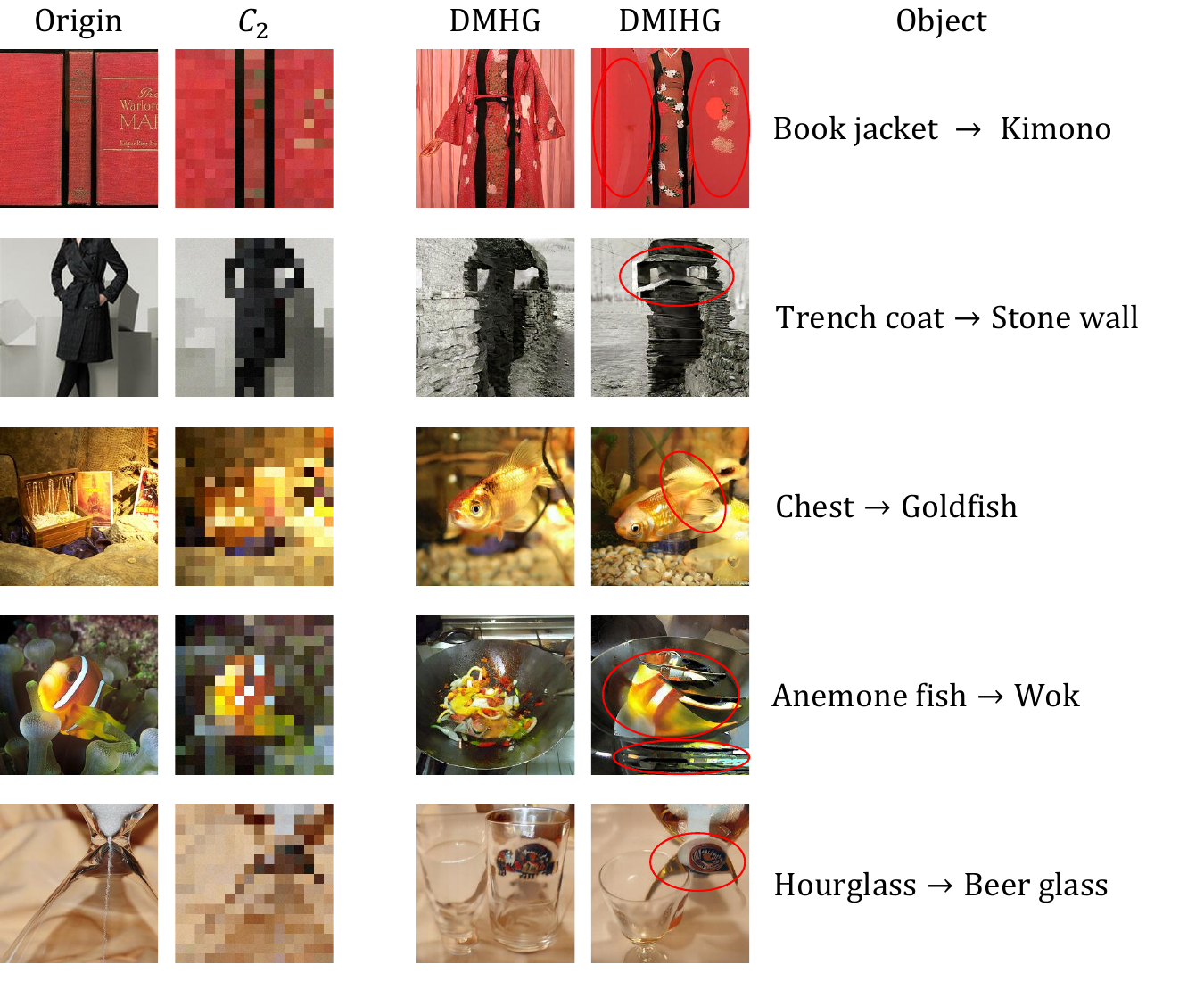}
    \caption{Extra examples of inpainting and object editing tasks. The images generated by DMIHG tend to exhibit some subtle illusions, which are highlighted with red circles. The guidance scales of inpainting are $\lambda_1 = 2.5$ and $\lambda_2 = 100$. And the guidance scales of object editing are $\lambda_1 = 4.5$ and $\lambda_2 = 200$}
    \label{fig_img_more_examples}
\end{figure*}

Figure \ref{fig_generated_mols_task12} showcases a selection of molecules that satisfy Task 1 and Task 2 (\ref{eq_molecular_task_12}) in Section \ref{sec_design_mole_from_0}, while Figure \ref{fig_partial_diff_all} present several successful examples of molecules generated through partial DMDG in Section \ref{sec_nearby_sampling}.  Some results of image inpainting and object editing have already been shown in Figures~\ref{fig_inpaint_obj_edit} and~\ref{fig_inpaint_edit}. In addition, we present some results in  Figure~\ref{fig_img_more_examples}, together with comparisons to DPS and DMIHG.

\end{document}